\documentclass[10pt,twocolumn,letterpaper]{article}
\usepackage[pagenumbers]{cvpr}
\usepackage{bm,mathtools}
\usepackage{multirow,array,tabularx,makecell}
\usepackage{microtype}
\usepackage{hyperref}
\usepackage[capitalize,noabbrev]{cleveref}
\usepackage{pifont}
\newcommand{\cmark}{\ding{51}}
\newcommand{\xmark}{\ding{55}}
\newcommand{\ours}{\textsc{GR-GIC}\xspace}

\newcommand{\subfigtarget}[1]{\hypertarget{#1}{}}
\makeatletter
\renewcommand{\p@subfigure}{\thefigure}

\newcommand{\subfiglabel}[1]{%
  \refstepcounter{subfigure}%
  \begingroup
    \protected@edef\@currentHref{#1}%
    \label{#1}%
  \endgroup
}
\makeatother

\title{Efficient Generative Image Compression with Quantized Latents Refinement}
\author{%
Hao Cao$^{1}$\quad
Yinhuan Huang$^{1}$\quad
Wenqi Guo$^{2}$\quad
Zhijin Qin$^{1,3}$\thanks{Corresponding author.}\quad
Jungong Han$^{2,4}$\\
$^{1}$Department of Electronic Engineering, Tsinghua University\\
$^{2}$Department of Automation, Tsinghua University\\
$^{3}$State Key Laboratory of Space Network and Communications\\
$^{4}$Beijing National Research Center for Information Science and Technology
}
\hypersetup{
  pdftitle={Efficient Generative Image Compression with Quantized Latents Refinement},
  pdfauthor={Hao Cao, Yinhuan Huang, Wenqi Guo, Zhijin Qin, Jungong Han}
}

\begin{document}
\maketitle

\begin{abstract}
Generative image compression (GIC) delivers strong perceptual quality at low bitrates.
However, GIC methods with strong perceptual quality often rely on pretrained diffusion models with high latency, while recent lightweight GIC methods reduce latency but deliver lower perceptual quality.
We observe that, for lightweight GIC methods at low bitrates, coarse deterministic quantization can collapse nearby encoder latents corresponding to different fine details to the same point, making these details difficult to recover.
Motivated by this, we propose \ours, an efficient GIC that delivers competitive reconstruction performance with fast inference.
For better performance, we introduce Guided Refinement (GR), which combines latent refinement guided by entropy context with adaptive perturbations for training regularization, jointly improving reconstruction at low bitrates.
For fast inference, we introduce reorganized projected residual vector quantization (P-RVQ), which equivalently reorganizes residual updates into low-dimensional lookups to expose greater parallelism and reduce sequential computation.
Experiments on Kodak show that for performance, compared with OneDC, \ours achieves BD-rate savings of 23.19\%, 37.71\%, 22.30\%, and 22.14\% for LPIPS, DISTS, PSNR, and FID, respectively.
For fast inference, the highly optimized implementation of \ours encodes and decodes in 4.60 and 5.98 ms on an NVIDIA A100.
Specifically, the reorganized P-RVQ implementation achieves a $3.14\times$ average speedup over the original implementation under the same FP32 setting.
On an NVIDIA Jetson Orin NX, \ours achieves latency comparable to lightweight GIC methods.
\end{abstract}
\section{Introduction}

\label{sec:intro}

\begin{figure*}[t]
  \centering
  \subfigtarget{fig:motivation_efficiency-a}%
  \subfigtarget{fig:motivation_efficiency-b}%
  \subfigtarget{fig:motivation_efficiency-c}%
  \subfigtarget{fig:motivation_efficiency-d}%
  \includegraphics[width=0.85\linewidth]{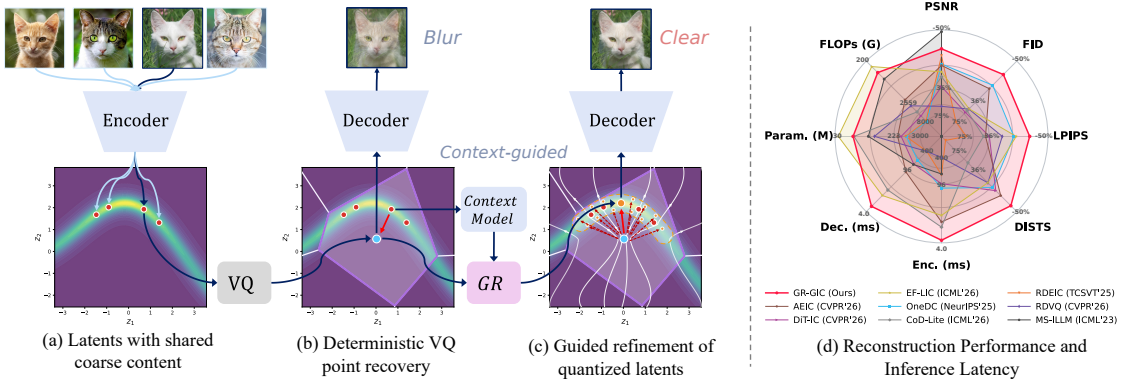}
  \caption{Motivation and overview of \ours.
  (a) Images with similar coarse content but different fine details are mapped to nearby points in the continuous latent space.
  (b) At low bitrates, coarse deterministic quantization can collapse these distinct latents to the same quantized point, leaving their fine detail differences unspecified.
  (c) GR uses entropy context to define a Gaussian refinement distribution around the recovered latent and introduces adaptive perturbations during training.
  (d) GR-GIC combines competitive reconstruction performance with fast inference.}
  \label{fig:motivation_efficiency}
  \setcounter{subfigure}{0}
  \subfiglabel{fig:motivation_efficiency-a}
  \subfiglabel{fig:motivation_efficiency-b}
  \subfiglabel{fig:motivation_efficiency-c}
  \subfiglabel{fig:motivation_efficiency-d}
\end{figure*}

Generative image compression (GIC) provides strong perceptual reconstruction at low bitrates.
This is challenging for traditional codecs and learned image codecs optimized primarily for pixel fidelity, which tend to smooth textures and weaken fine details as the bitrate decreases~\citep{jpeg,vtm,lic-balle2018variational,hpcm}.
GIC addresses this limitation with adversarial generators~\citep{hific,ms-illm}, discrete generative representations~\citep{vq-gic,vq-DLF}, or diffusion priors~\citep{diffusion-perco,diffusion-StableCodec,diffusion-xue2025one}, whose learned priors can synthesize plausible details that are not fully specified by the compressed representation.

However, high-performing GIC methods often rely on pretrained diffusion models, whose large parameter counts and FLOPs hinder practical deployment~\citep{oscar,diffusion-StableCodec,diffusion-xue2025one,ditic2026}.
To address this issue, recent work has explored lightweight GIC methods with fewer parameters and faster inference~\citep{progic2026,efic2026,codlite2026}, but their reconstruction performance still leaves room for improvement compared with diffusion-based methods.
This leaves a trade-off between reconstruction performance and fast inference.

We observe that, in lightweight GIC methods, images with similar coarse content but different fine details can be mapped to nearby points in the continuous latent space.
At low bitrates, coarse deterministic quantization can collapse these distinct latents to the same quantized point, making their fine detail differences unavailable from the quantized representation, as illustrated in \cref{fig:motivation_efficiency-a,fig:motivation_efficiency-b}.
By contrast, diffusion based methods introduce diversity through Gaussian noise and use strong generative priors to synthesize plausible details, providing a natural mechanism to compensate for this uncertainty.

Motivated by this observation, we propose \ours, an efficient GIC that combines competitive reconstruction performance with fast inference, as shown in \cref{fig:motivation_efficiency-d}.
For better performance, we introduce Guided Refinement (GR), which uses entropy context to parameterize a Gaussian refinement distribution around the recovered latent.
Its mean shifts the recovered latent toward a refined center, while its variance controls adaptive perturbations for training regularization, jointly improving reconstruction at low bitrates.
For fast inference, we target projected residual vector quantization (P-RVQ), which is a major encoder latency bottleneck in EF-LIC~\citep{efic2026}.
We derive an algebraically equivalent reorganization that replaces serial high dimensional residual updates with low dimensional lookups, exposing greater parallelism and reducing sequential computation.

Experiments on Kodak show that (i) for performance, compared with OneDC~\citep{diffusion-xue2025one}, \ours achieves BD-rate~\citep{bd-rate} savings of 23.19\%, 37.71\%, 22.30\%, and 22.14\% for LPIPS~\citep{lpips}, DISTS~\citep{dists}, PSNR, and FID~\citep{fid}, respectively.
(ii) For fast inference, the highly optimized implementation of \ours encodes and decodes in 4.60 and 5.98 ms on an NVIDIA A100, achieving $16.73\times$ and $34.19\times$ speedups over the official OneDC implementation, respectively.
Specifically, the reorganized P-RVQ implementation achieves a $3.14\times$ average speedup over the original implementation under the same FP32 setting.
We further validate \ours on an NVIDIA Jetson Orin NX, where it achieves latency comparable to existing lightweight GIC methods, supporting its deployment on edge devices.

Our contributions are summarized as follows:

\begin{itemize}
\item We introduce \ours, an efficient GIC that combines GR and reorganized P-RVQ for competitive reconstruction performance and fast inference.
\item GR uses entropy context to parameterize an adaptive refinement distribution around quantized latents. Reorganized P-RVQ replaces serial high dimensional residual updates with low dimensional lookups, reducing sequential computation and inference latency.
\item For performance, experiments demonstrate competitive results across multiple reconstruction metrics.
For fast inference, the highly optimized implementation achieves low latency on both GPU and edge platforms.
\end{itemize}

\section{Related Work}
\label{sec:related}

\paragraph{Diffusion-based generative image compression.}
Diffusion models provide strong generative priors for reconstructing details that are not fully specified by compressed representations~\citep{diffusion-cdc,diffusion-perco}.
Recent GIC methods increasingly use pretrained diffusion models and improve their efficiency through fewer diffusion steps, one-step reconstruction, or lighter architectures~\citep{rdeic,diffusion-StableCodec,oscar,diffusion-xue2025one,codlite2026}.
Despite these advances, their model size and computation remain substantial for efficient deployment.

\paragraph{VAE-based generative image compression.}
VAE-based GIC provides a more efficient alternative through adversarial priors, improved latent representations, and vector quantization~\citep{hific,ms-illm,glc2024,cgic,vq-DLF,rdvq2026}.
Recent work further develops lightweight GIC.
ProGIC combines a depthwise convolutional backbone with RVQ~\citep{progic2026}, while EF-LIC models correlations among latents in the representation domain~\citep{efic2026}.
\ours builds on EF-LIC, introducing GR for better reconstruction and reorganizing P-RVQ to move high dimensional reconstruction outside the sequential search path.

\paragraph{Latent refinement after quantization.}
Prior LIC methods have compensated quantization loss through latent residual prediction, post-quantization filtering, and adaptive quantization or correction~\citep{minnen2020channel,fu2023asymmetric,zhang2024another,ge2024nlic}.
Recent work also studies latent and noise compensation~\citep{yi2026rethinking}.
These methods mainly perform deterministic correction using decoder context, rely on guidance from the encoder for content-adaptive compensation, or model stochastic compensation without adapting it to context.
In contrast, GR uses entropy context already available at the decoder to parameterize a refinement distribution that adapts to spatial location and image content, without additional payload, while adaptive perturbations provide training regularization for generative reconstruction.

\section{Method}
\label{sec:method}

\begin{figure*}[t]
  \centering
  \includegraphics[width=0.85\textwidth]{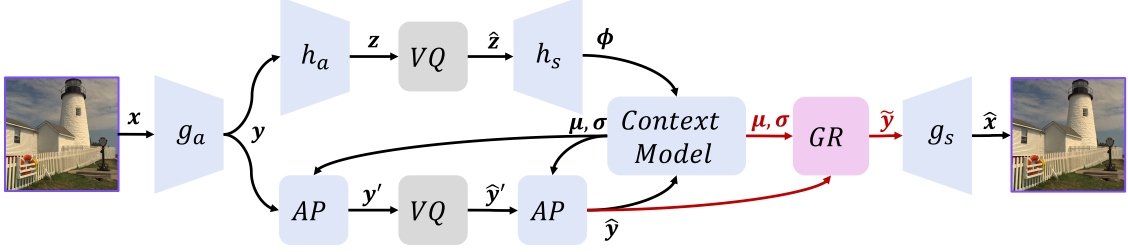}
  \caption{Overview of \ours.
The encoder $g_a$ produces the latent $\bm y$, while the hyper encoder $h_a$ and hyper decoder $h_s$ provide the context feature $\bm\phi$.
The context model predicts $(\bm\mu,\bm\sigma)$ for the groupwise affine projections (APs) and P-RVQ.
The GR head refines the recovered latent $\hat{\bm y}$ into $\tilde{\bm y}$ before the decoder $g_s$ reconstructs $\hat{\bm x}$.}
  \label{fig:framework}
\end{figure*}

\subsection{Overview}
\label{sec:overview}
\begin{figure*}[t]
  \centering
  \subfigtarget{fig:rvq_reorganization-a}%
  \subfigtarget{fig:rvq_reorganization-b}%
  \includegraphics[width=\textwidth]{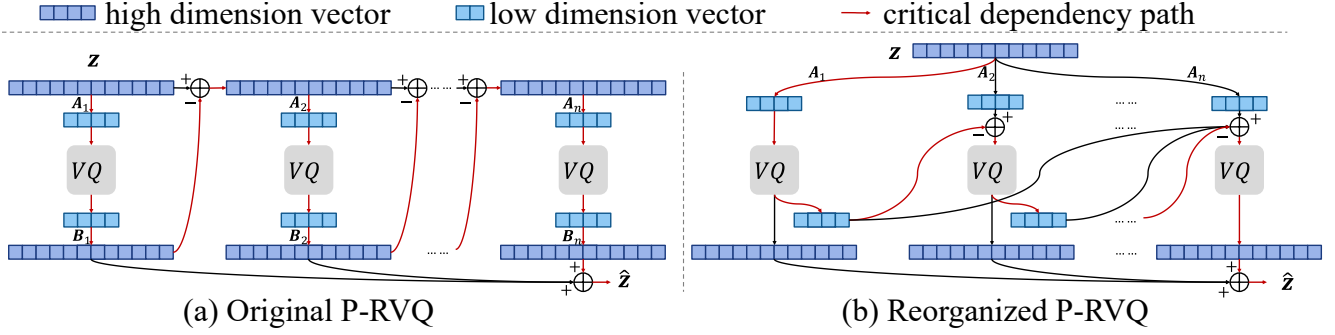}
  \caption{Reorganized P-RVQ.
(a) Original P-RVQ alternates high dimensional residual updates, projections, and greedy search.
(b) The equivalent form stacks source projections and uses precomputed low dimensional lookups.
  Red arrows mark the critical dependency path.}
  \label{fig:rvq_reorganization}
  \setcounter{subfigure}{0}
  \subfiglabel{fig:rvq_reorganization-a}
  \subfiglabel{fig:rvq_reorganization-b}
\end{figure*}

As illustrated in \cref{fig:motivation_efficiency-a,fig:motivation_efficiency-b}, at low bitrates, coarse deterministic quantization can collapse nearby encoder latents corresponding to different fine details to the same quantized point, making these details difficult to recover in lightweight GIC.
GR-GIC builds on EF-LIC~\citep{efic2026}, whose context model provides the recovered latent and entropy context without additional payload.
GR uses the entropy context to parameterize a Gaussian refinement distribution around the recovered latent, with adaptive perturbations providing latent noise regularization during training.
GR-GIC retains the analysis transform, hyperprior, groupwise context, and P-RVQ structure of EF-LIC, adds a GR head before synthesis as shown in \cref{fig:framework}, and reorganizes P-RVQ execution in \cref{fig:rvq_reorganization}.

Given an image $\bm x\in\mathbb R^{3\times H\times W}$, the encoder $g_a$ produces the latent $\bm y$.
The hyper encoder $h_a$ maps $\bm y$ to the hyper latent $\bm z$, which the hyper latent RVQ $Q_{\bm z}$ quantizes to obtain $\hat{\bm z}$.
The hyper decoder $h_s$ reconstructs the context feature $\bm\phi$ from $\hat{\bm z}$:
\[
  \begin{aligned}
    \bm y&=g_a(\bm x), & \bm z&=h_a(\bm y),\\
    \hat{\bm z}&=Q_{\bm z}(\bm z), & \bm\phi&=h_s(\hat{\bm z}).
  \end{aligned}
\]
The latent $\bm y$ has a spatial downsampling factor of $f_y=16$ and enters the groupwise affine projections.
Following the quadtree context modeling of EF-LIC and DCVC-RT~\citep{efic2026,dcvc-rt}, we partition $\bm y$ into $N=4$ groups $\{\bm y_n\}_{n=1}^{N}$.
For group $n$, the context model $f_n$ combines $\bm\phi$ with the previously recovered groups to predict the mean $\bm\mu_n$ and scale $\bm\sigma_n$:
\[
  \begin{gathered}
  (\bm\mu_n,\bm\sigma_n)
  =f_n\!\left(\operatorname{concat}(\bm\phi,\hat{\bm y}_{<n})\right),\\
  \bm\sigma_n\geq\sigma_{\min}.
  \end{gathered}
\]
Before quantization, the affine projection (AP) centers and scales $\bm y_n$ using $\bm\mu_n$ and $\bm\sigma_n$.
The inverse AP then reconstructs $\hat{\bm y}_n$ after quantization:
\begin{equation}
  \begin{aligned}
    \bm y'_n&=(\bm y_n-\bm\mu_n)\oslash\bm\sigma_n,\\
    \hat{\bm y}'_n&=Q_n(\bm y'_n),\\
    \hat{\bm y}_n&=\bm\sigma_n\odot\hat{\bm y}'_n+\bm\mu_n.
  \end{aligned}
  \label{eq:conditional_quantization}
\end{equation}
After all groups are recovered, their concatenation forms $\hat{\bm y}$.
The recovered latent $\hat{\bm y}$ and the aligned context parameters $(\bm\mu,\bm\sigma)$ are then fed to the GR head $r_\theta$ to produce $\tilde{\bm y}$, which the decoder $g_s$ maps to the reconstruction $\hat{\bm x}=g_s(\tilde{\bm y})$.

The codec uses five RVQs, with $Q_{\bm z}$ quantizing the hyper latent $\bm z$ and $Q_n$ quantizing each of the $N=4$ main latent groups.
At a given operating point, all five RVQs activate the same number $M\in\{1,2,3,4\}$ of codebooks, and $m\in\{1,\ldots,M\}$ indexes an active stage.
With this shared RVQ depth and fixed width indices, the bit rate is $R_M=0.0356M$ bits per pixel (BPP).
\cref{app:nominal_rate,app:padding} provide the exact rate calculation and boundary handling.

\subsection{Guided Refinement of Quantized Latents}
\label{sec:gr}

To improve low-bitrate reconstruction in lightweight GIC, we introduce Guided Refinement (GR) after RVQ recovery.
GR uses entropy context to define a Gaussian refinement distribution around the recovered latent that adapts to spatial location and image content.
During joint training, sampling from this distribution introduces adaptive perturbations that provide latent noise regularization.

We first formalize the effect of deterministic quantization that motivates this refinement.
To express the mapping with the notation of \cref{eq:conditional_quantization}, we suppress group and stage indices, write $Q$ for the deterministic quantizer, and denote its $j$th reconstruction point by $\hat{\bm y}'_j$.
Its quantization cell is $\mathcal C_j=\{\bm y':Q(\bm y')=\hat{\bm y}'_j\}$, and inverse projection gives $\hat{\bm y}_j=\bm\sigma\odot\hat{\bm y}'_j+\bm\mu$.
The continuous latent distribution is transformed as

\begin{equation}
  \begin{aligned}
  &p_{\hat{\bm Y}}(\hat{\bm y}\mid\bm\mu,\bm\sigma)\\
  &\quad=\int p_{\bm Y'}(\bm y'\mid\bm\mu,\bm\sigma)\\
  &\qquad\quad\times p_{\hat{\bm Y}}(\hat{\bm y}\mid\bm Y'=\bm y',\bm\mu,\bm\sigma)
  \,\mathrm d\bm y'\\
  &\quad=\sum_j\left(
      \int_{\mathcal C_j}
      p_{\bm Y'}(\bm y'\mid\bm\mu,\bm\sigma)\,\mathrm d\bm y'
    \right)\\
  &\qquad\quad\times\delta_{\mathrm D}(\hat{\bm y}-\hat{\bm y}_j).
  \end{aligned}
  \label{eq:conditional_atomic_distribution}
\end{equation}
Here $\delta_{\mathrm D}$ is the Dirac delta.
\Cref{eq:conditional_atomic_distribution} shows that deterministic quantization assigns all latent realizations within one quantization cell to the same recovered point.
The transmitted index therefore identifies the quantization cell, but does not preserve the continuous variation within that cell.
At low bitrates, where the quantization cells are relatively coarse, this unresolved variation can correspond to different fine details that are difficult to determine from the recovered latent alone.
This motivates learning a refinement distribution around the recovered point using information already provided by the entropy context.

We therefore use a Gaussian family conditioned on entropy context, as its simple mean-variance form separates the refinement center from adaptive perturbations without introducing additional latent correlations.
Concatenating the groupwise reconstruction points in \cref{eq:conditional_atomic_distribution} gives the recovered latent $\hat{\bm y}$.
Let $\bm\epsilon\sim\mathcal N(\bm0,\bm I)$ denote a standard Gaussian variable for one latent realization.
We model the refined latent distribution as
\begin{equation}
  \begin{aligned}
  &p_\theta(\tilde{\bm y}\mid\hat{\bm y},\bm\mu,\bm\sigma)\\
  &\quad=\int p_{\bm\epsilon}(\bm\epsilon)\,
  p_\theta\bigl(
    \tilde{\bm y}\mid
    \bm\epsilon,\hat{\bm y},\bm\mu,\bm\sigma
  \bigr)\,\mathrm d\bm\epsilon\\
  &\quad=\mathcal N\!\left(
    \tilde{\bm y}\mid
    \hat{\bm y}+\bm\delta,
    \operatorname{diag}(\bm\beta^2)
  \right).
  \end{aligned}
  \label{eq:gr_conditional_distribution}
\end{equation}
Here $\bm\delta$ shifts the recovered latent to the refinement center $\hat{\bm y}+\bm\delta$, while $\bm\beta$ controls the perturbation scale across spatial locations and image content.
Both quantities are predicted from the recovered latent and entropy context parameters that are already available after decoding:
\begin{equation}
  \begin{gathered}
  (\bm\delta,\tilde{\bm\beta})
  =
  r_\theta\!\left(
    \operatorname{concat}(\hat{\bm y},\bm\mu,\bm\sigma)
  \right),\\
  \bm\beta=\max(\tilde{\bm\beta},\bm0).
  \end{gathered}
  \label{eq:gr_parameters}
\end{equation}
The refined latent is then sampled as
\begin{equation}
  \bm\epsilon\sim\mathcal N(\bm0,\bm I),
  \qquad
  \tilde{\bm y}
  =
  \hat{\bm y}
  +
  \bm\delta
  +
  \bm\beta\odot\bm\epsilon.
  \label{eq:gr_sample}
\end{equation}
All codec components are jointly optimized with one independent Gaussian draw per crop.
The learned refinement center adjusts the recovered latent according to the entropy context, while the adaptive perturbation provides latent noise regularization during training.
\Cref{sec:ablation} further supports the important role of $\bm\beta\odot\bm\epsilon$ during training.

\subsection{Reorganized Projected Residual Vector Quantization}
\label{sec:rvq_optimization}
EF-LIC~\citep{efic2026} reported that P-RVQ takes 9.16 ms on Kodak, accounting for 52.0\% of the measured 17.62 ms encoder latency and forming a major encoder latency bottleneck.
For fast inference, we reorganize P-RVQ execution while preserving its greedy quantization process.

Within this subsection, $\bm z\in\mathbb R^D$ denotes an abstract input to a hyper or main latent P-RVQ.
At stage $m$, $\bm A_m\in\mathbb R^{d\times D}$ projects the residual from high to low dimensions for search.
The codebook $\bm E_m=[\bm E_m[1],\ldots,\bm E_m[K]]\in\mathbb R^{d\times K}$ stores $K$ low dimensional codewords as columns, and $\bm B_m\in\mathbb R^{D\times d}$ maps them back to the reconstruction space, with projection biases suppressed.
Starting from $\hat{\bm z}_0=\bm 0$, the partial reconstruction is defined for $m=1,\ldots,M$ and returns $\hat{\bm z}=\hat{\bm z}_M$.
As illustrated in \cref{fig:rvq_reorganization-a}, the original recursion carries a state in $\mathbb R^D$, projects the residual into $\mathbb R^d$ for search, and maps the selected codeword back to $\mathbb R^D$ before the next stage, so both projection directions remain on the critical dependency path:
\begin{equation}
  \begin{gathered}
  \textsc{Original:}\\
  \hat{\bm z}_m
  =\underbrace{
    \begin{gathered}
      \hat{\bm z}_{m-1}
      +\bm B_m\bm E_m\!\Bigl[\\
        \arg\min_{1\leq j\leq K}
        \left\|
          \bm A_m(\bm z-\hat{\bm z}_{m-1})
          -\bm E_m[j]
        \right\|_2^2
      \Bigr]
    \end{gathered}
    }_{\substack{\text{critical dependency path with high dimensional updates:}\\\mathbb R^D\rightarrow\mathbb R^d\rightarrow\mathbb R^D}}.
  \end{gathered}
  \label{eq:projected_rvq_original}
\end{equation}

We derive an algebraically equivalent reorganization that stacks the source projections and expresses subsequent residual updates as low dimensional lookups, as illustrated in \cref{fig:rvq_reorganization-b}.
Since the projection matrices and codebooks are fixed after training, each product
$\bm A_m\bm B_\ell\bm E_\ell\in\mathbb R^{d\times K}$
can be fused offline into a new $d$ dimensional codebook.
Here, $i_\ell$ denotes the codeword index selected at an earlier stage $\ell$:
\begin{equation}
  \begin{gathered}
  \textsc{Reorganized:}\\
  \hat{\bm z}_m=\hat{\bm z}_{m-1}+\bm B_m\bm E_m\Biggl[\\
      \arg\min_{1\leq j\leq K}\Biggl\|
        \bm A_m\bm z
        -\underbrace{
          \sum_{\ell<m}
          \bm A_m\bm B_\ell\bm E_\ell[i_\ell]
        }_{\mathclap{\substack{\text{critical dependency path using new}\\d\text{ dimensional codebooks fused offline}}}}
        -\bm E_m[j]
      \Biggr\|_2^2
    \Biggr].
  \end{gathered}
  \label{eq:projected_rvq_rewrite}
\end{equation}
The reorganized search evaluates all source projections together and updates later search states through lookups in $\mathbb R^d$.
The high dimensional reconstruction is gathered from offline tables only after the codeword indices are selected.
By linearity, the original and reorganized forms preserve the same search states, greedy assignments, and final reconstruction in real arithmetic.
The reorganization therefore moves high dimensional reconstruction outside the critical dependency path, exposing greater parallelism and shortening the sequential computation path.

Under the same Torch FP32 backend and precision, the reorganized P-RVQ implementation achieves a $3.14\times$ average speedup over the original implementation across the 16 evaluated stage count and resolution settings.
This matched comparison isolates the gain from the algebraic reorganization, which accounts for $23.54\%$ of the total encoder latency reduction.
The CUDA FP16 implementation further increases the average speedup to $7.15\times$ relative to the original FP32 implementation.
At the highest evaluated Kodak BPP, compression-only P-RVQ reaches an $18.80\times$ speedup, and the complete P-RVQ acceleration accounts for $31.28\%$ of the total encoder latency reduction.

\section{Experiments}
\label{sec:experiments}

\subsection{Experimental Setup}

\paragraph{Implementation Details.}
Our implementation largely follows EF-LIC~\citep{efic2026}, with a GR head added before synthesis and P-RVQ execution reorganized as described above.
\Cref{app:details} provides the complete architecture and codebook settings.
For fast inference, the highly optimized implementation further uses fused CUDA operators following DCVC-UF~\citep{dcvc-uf}.

\paragraph{Training Details.}
Each operating point is trained on ImageNet~\citep{imagenet} in two 1M-iteration stages.
Stage one samples 1\% per epoch with $256\!\times\!256$ random crops at a learning rate of $10^{-4}$.
Stage two samples 10\% of images exceeding 512 pixels in both dimensions with $512\!\times\!512$ random crops at $10^{-5}$.
Both stages apply random horizontal flipping and use AdamW~\citep{adamw} with weight decay $10^{-4}$, $(\beta_1,\beta_2)=(0.9,0.999)$, batch size 16, and one NVIDIA A100.
Each training crop uses one standard Gaussian draw $\bm\epsilon\sim\mathcal N(\bm0,\bm I)$ to sample the refinement distribution conditioned on entropy context and estimate the expected image-space objective, which includes a generative adversarial network (GAN) term:
\begin{equation}
  \begin{aligned}
  \mathcal L={}&\mathbb E_{\bm\epsilon}\!\bigl[\lambda_{\mathrm{rec}}\mathcal L_2
    +\lambda_{\mathrm{per}}\mathcal L_{\mathrm{LPIPS}}
    +\lambda_{\mathrm{adv}}\mathcal L_{\mathrm{GAN}}\bigr]\\
    &+\lambda_{\mathrm{cb}}\mathcal L_{\mathrm{cb}}
    +\lambda_{\mathrm{com}}\mathcal L_{\mathrm{com}}.
  \end{aligned}
  \label{eq:training_objective}
\end{equation}
Here $\mathcal L_2=\|\bm x-\hat{\bm x}\|_2^2$ is averaged over pixels and channels, $\mathcal L_{\mathrm{GAN}}$ is the generator-side PatchGAN term, and $\mathcal L_{\mathrm{cb}}$ and $\mathcal L_{\mathrm{com}}$ are the codebook and commitment terms.
We use $\lambda_{\mathrm{rec}}=18$, $\lambda_{\mathrm{per}}=2$, $\lambda_{\mathrm{cb}}=1$, and $\lambda_{\mathrm{com}}=1$, with adaptive $\lambda_{\mathrm{adv}}$ following VQGAN~\citep{vq-gan}.
All configurations in the study share the PatchGAN rule, training data, and losses.
For the final codec, the adaptive perturbation provides latent noise regularization with all codec components jointly optimized.

\paragraph{Evaluation Datasets and Metrics.}
We evaluate Kodak, Tecnick, DIV2K validation, and CLIC 2020 Professional~\citep{kodak,tecnick,div2k,clic2020}.
Metrics are LPIPS with a Visual Geometry Group (VGG) backbone, DISTS, PSNR, and patch-based FID~\citep{lpips,vgg,dists,fid,glc2024}.
LPIPS and DISTS capture perceptual similarity, PSNR reports RGB pixel fidelity, and FID measures distribution-level alignment.
All codecs use the same metric pipeline on decoded images for direct curves and pairwise BD-rate calculations.
Appendix~\ref{app:protocol} gives dataset, metric, and pairwise PCHIP details~\citep{bd-rate}.

\begin{figure*}[t]
  \centering
  \includegraphics[width=0.82\linewidth]{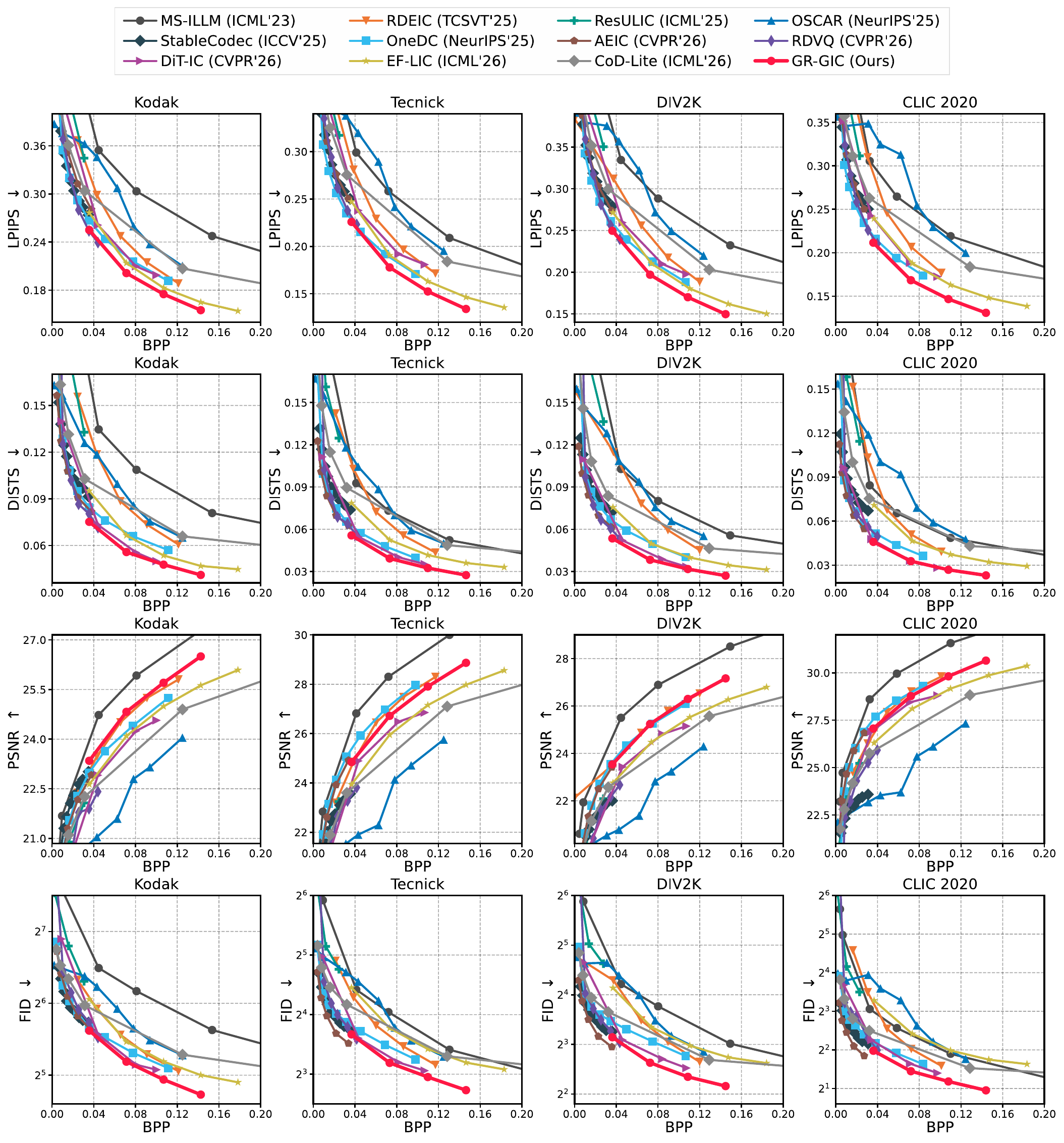}
  \caption{Rate and reconstruction quality across four datasets.
Lower LPIPS, DISTS, and FID and higher PSNR indicate better quality.
FID uses base-2 logarithmic vertical axes.}
  \label{fig:rd_curves}
\end{figure*}

\paragraph{Compared Methods.}
The pretrained diffusion group comprises ResULIC, StableCodec, RDEIC, OneDC, OSCAR, DiT-IC, AEIC, and CoD-Lite~\citep{resulic2025,diffusion-StableCodec,rdeic,diffusion-xue2025one,oscar,ditic2026,aeic2026,codlite2026}.
The VAE-based group comprises MS-ILLM, Control-GIC, GLC, DLF, ProGIC, RDVQ, EF-LIC, and \ours~\citep{ms-illm,cgic,glc2024,vq-DLF,progic2026,rdvq2026,efic2026}.
Runtime uses batch size one with $512\!\times\!768$ images on one NVIDIA A100.
Latency follows the implementation and timing protocols in \cref{app:runtime}, with parameters and forward FLOPs reported as additional model complexity measures.
CoD-Lite uses our measurements of its official code, with its paper profile in \cref{app:baseline_protocol}.

\subsection{Comparison with Previous Methods}
\label{sec:comparison}

\begin{table*}[t]
  \centering
  \caption{Results on the Kodak dataset, with latency measured on one NVIDIA A100 GPU.
Encoding and decoding latencies are in milliseconds (ms), FLOPs in billions (G), parameters in millions (M), and BD-rate in percent.
OneDC is the BD-rate anchor, and lower values are better for every reported measure.
Bold and underlined values indicate the best and second best results, respectively.}
  \label{tab:main_results}
  \begingroup
  \setlength{\tabcolsep}{3.45pt}
  \begin{tabular}{@{}llrrrrrrrr@{}}
    \toprule
    \multirow{2}{*}{Family} & \multirow{2}{*}{Method}
      & \multicolumn{4}{c}{Complexity}
      & \multicolumn{4}{c}{BD-rate (\%)} \\
    \cmidrule(lr){3-6}
    \cmidrule(l){7-10}
      & & Enc.
& Dec.
& FLOPs & Params
      & LPIPS & DISTS & PSNR & FID \\
    \midrule
    \multirow{8}{*}{\makecell{Pretrained\\diffusion}} & ResULIC
      & 96.24 & 428.32 & 8549.01 & 1390.00
      & 206.40 & 260.74 & 10.77 & 304.92 \\
    & StableCodec
      & 100.18 & 200.03 & 6820.28 & 1065.81
      & 8.03 & 6.77 & -17.10 & \underline{-4.45} \\
    & RDEIC
      & 157.25 & 426.68 & 7767.46 & 1380.27
      & 69.79 & 166.61 & -12.14 & 80.26 \\
    & OneDC
      & 76.98 & 204.47 & 8557.18 & 1406.42
      & 0.00 & 0.00 & 0.00 & 0.00 \\
    & OSCAR
      & 53.04 & 167.56 & 6485.61 & 1009.30
      & 127.15 & 131.50 & 154.09 & 159.92 \\
    & DiT-IC
      & 97.62 & 151.89 & 5743.15 & 1069.41
      & 35.99 & -6.44 & 33.24 & 56.47 \\
    & AEIC
      & 12.39 & 82.98 & 2143.06 & 885.70
      & 41.65 & \underline{-16.71} & 2.11 & 7.16 \\
    & CoD-Lite
      & 9.33 & 20.86 & 771.69 & 80.15
      & 44.53 & 51.35 & 18.72 & 53.06 \\
    \midrule
    \multirow{8}{*}{VAE-based} & MS-ILLM
      & 165.38 & 147.79 & 599.52 & 181.40
      & 293.59 & 364.46 & \textbf{-47.46} & 468.10 \\
    & Control-GIC
      & 103.56 & 436.26 & 5816.37 & 130.36
      & 602.72 & 807.48 & 175.78 & 855.14 \\
    & GLC
      & 34.75 & 48.26 & 2466.28 & 163.99
      & 21.53 & 54.70 & 56.04 & 82.95 \\
    & DLF
      & 168.78 & 240.63 & 5880.53 & 1286.01
      & 20.87 & 20.19 & 42.73 & 70.80 \\
    & ProGIC
      & \underline{7.64} & \underline{10.99} & \underline{333.38} & \textbf{33.11}
      & 36.47 & 80.41 & 49.84 & 156.32 \\
    & RDVQ
      & 408.24 & 412.44 & 3090.29 & 251.84
      & 17.06 & 9.87 & 62.13 & 49.51 \\
    & EF-LIC
      & 17.62 & 13.72 & \textbf{279.61} & \underline{35.74}
      & \underline{-3.33} & 10.05 & 11.49 & 49.01 \\
    \cmidrule(l){2-10}
    & \textbf{\ours}
      & \textbf{4.60} & \textbf{5.98} & 456.55 & 79.39
      & \textbf{-23.19} & \textbf{-37.71}
      & \underline{-22.30} & \textbf{-22.14} \\
    \bottomrule
  \end{tabular}
  \endgroup
\end{table*}

\Cref{fig:rd_curves,tab:main_results} compare the reconstruction performance and inference latency of \ours with previous GIC methods.
For performance, \ours achieves bitrate savings over OneDC for LPIPS, DISTS, and FID across all four datasets.
On Kodak, the OneDC-anchored BD-rate savings are 23.19\%, 37.71\%, 22.30\%, and 22.14\% for LPIPS, DISTS, PSNR, and FID, respectively.

For fast inference, the highly optimized implementation of \ours encodes and decodes a $512\times768$ image in 4.60 and 5.98 ms on an NVIDIA A100.
The official OneDC implementation takes 76.98 and 204.47 ms under the evaluated system settings, making the corresponding latency ratios $16.73\times$ and $34.19\times$.
\Cref{app:runtime} details the numerical precision, acceleration paths, and timing boundaries used for the compared systems.

Among the evaluated VAE-based methods, \ours improves the rate-quality curves of previous lightweight codecs across the reported reconstruction metrics while retaining low inference latency.
Taken together, these results show that \ours combines gains across multiple reconstruction metrics with fast inference.

\begin{figure*}[t]
  \centering
  \includegraphics[width=0.90\linewidth]{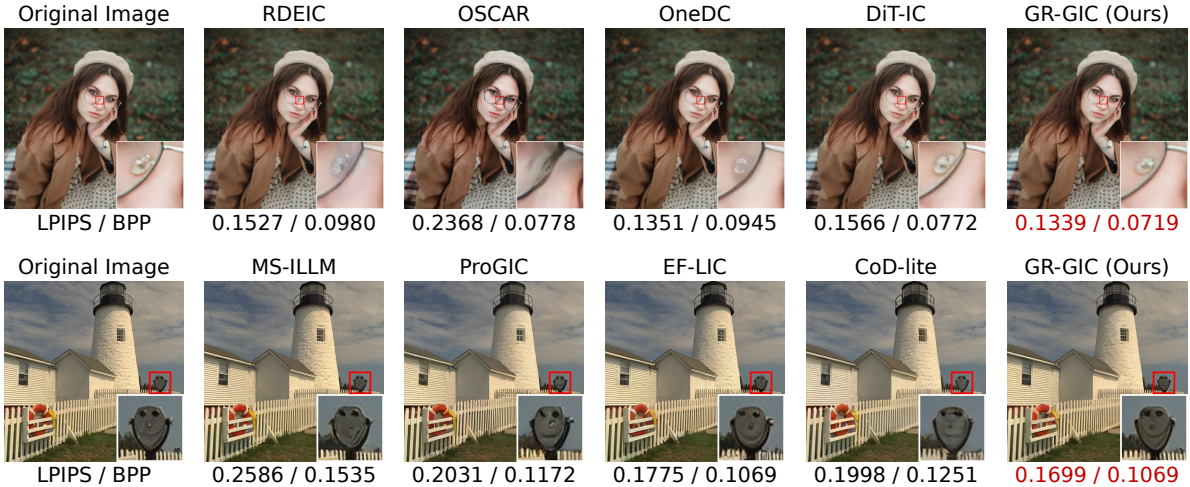}
  \caption{Qualitative comparisons of \ours with high-capacity diffusion and lightweight codecs at similar rates.
  Boxes mark local details.
  Labels report LPIPS and BPP, where lower is better.}
  \label{fig:qualitative_comparison}
\end{figure*}

At similar rates, the qualitative comparisons in \cref{fig:qualitative_comparison} further illustrate the reconstruction differences.
In the first row, \ours preserves the thin eyeglass frame and nose pad.
In the second row, it reconstructs the small circular fixture with coherent local geometry.
These examples are consistent with the quantitative gains and show clearer reconstruction of local structures.

\subsection{Guided Refinement Configuration Study}
\label{sec:ablation}

\begin{table*}[t]
  \centering
\caption{Guided Refinement configurations on Kodak at four nominal rates.
The GR columns indicate a learned offset, decoder context, stochastic scale, and adaptive scale.}
  \label{tab:ablation}
  \begingroup
  \setlength{\tabcolsep}{3.25pt}
  \begin{tabular}{lccccrrrrrrrr}
    \toprule
    \multirow{2}{*}{Model}
      & \multicolumn{4}{c}{GR configuration}
      & \multicolumn{2}{c}{BPP $=0.036$}
      & \multicolumn{2}{c}{BPP $=0.071$}
      & \multicolumn{2}{c}{BPP $=0.107$}
      & \multicolumn{2}{c}{BPP $=0.143$} \\
    \cmidrule(lr){2-5}
    \cmidrule(lr){6-7}
    \cmidrule(lr){8-9}
    \cmidrule(lr){10-11}
    \cmidrule(l){12-13}
      & Offset & Ctx.
& Stoch. & Adapt.
      & LPIPS & FID
      & LPIPS & FID
      & LPIPS & FID
      & LPIPS & FID \\
    \midrule
    $\mathrm{M}_0$
      & \xmark & \xmark & \xmark & \xmark
      & 0.277 & 66.2
      & 0.214 & 51.2
      & 0.184 & 43.6
      & 0.165 & 38.1 \\
    $\mathrm{M}_1$
      & \cmark & \xmark & \xmark & \xmark
      & 0.271 & 63.6
      & 0.212 & 46.8
      & 0.184 & 39.2
      & 0.164 & 35.4 \\
    $\mathrm{M}_2$
      & \cmark & \cmark & \xmark & \xmark
      & 0.264 & 62.4
      & 0.208 & 41.2
      & 0.182 & 36.5
      & 0.162 & 31.9 \\
    $\mathrm{M}_3$
      & \cmark & \cmark & \cmark & \xmark
      & 0.264 & 59.5
      & 0.207 & 37.9
      & 0.182 & 33.8
      & 0.162 & 29.5 \\
    \textbf{$\mathrm{M}_4$}
      & \cmark & \cmark & \cmark & \cmark
      & \textbf{0.261} & \textbf{54.6}
      & \textbf{0.206} & \textbf{36.0}
      & \textbf{0.180} & \textbf{32.3}
      & \textbf{0.160} & \textbf{27.6} \\
    \bottomrule
  \end{tabular}
  \endgroup
\end{table*}

\begin{table*}[t]
\centering
\caption{Encoding (Enc.) and decoding (Dec.) latency (ms) on an NVIDIA Jetson Orin NX.
Bold and underline mark the two best reported values in each column.}
\label{tab:edge_devices}
\begingroup
\setlength{\tabcolsep}{4pt}
\begin{tabular}{lrrrrrrrr}
\toprule
\multirow{2}{*}{Method} & \multicolumn{2}{c}{$512\!\times\!768$} & \multicolumn{2}{c}{1080p} & \multicolumn{2}{c}{2K} & \multicolumn{2}{c}{4K} \\
\cmidrule(lr){2-3}\cmidrule(lr){4-5}\cmidrule(lr){6-7}\cmidrule(l){8-9}
 & Enc.
& Dec.
& Enc.
& Dec.
& Enc.
& Dec.
& Enc.
& Dec.
\\
\midrule
EF-LIC & \textbf{71.253} & \underline{143.040} & \underline{347.490} & \underline{633.721} & \underline{623.651} & \underline{1137.985} & \underline{1441.824} & \underline{2628.537} \\
ProGIC & \underline{83.204} & 164.517 & 413.956 & 781.443 & 748.899 & 1412.819 & 1670.957 & 3182.561 \\
AEIC & 96.784 & 1290.483 & -- & {$>10$ s} & -- & {$>10$ s} & \multicolumn{2}{c}{OOM} \\
CoD-Lite & 114.456 & 239.371 & 564.422 & 1348.999 & 974.139 & 2489.412 & -- & {$>10$ s} \\
OneDC & 773.388 & 1981.831 & \multicolumn{2}{c}{OOM} & \multicolumn{2}{c}{OOM} & \multicolumn{2}{c}{OOM} \\
\midrule
\textbf{\ours} & 90.342 & \textbf{99.850} & \textbf{265.745} & \textbf{428.361} & \textbf{465.576} & \textbf{754.362} & \textbf{1061.516} & \textbf{1739.863} \\
\bottomrule
\end{tabular}
\endgroup
\end{table*}

We compare five refinement configurations, each trained from scratch for 1M iterations at each rate under identical optimization conditions.
$\mathrm{M}_0$ bypasses refinement.
$\mathrm{M}_1$ predicts an offset from the recovered latent, while $\mathrm{M}_2$ additionally uses entropy context.
$\mathrm{M}_3$ introduces stochastic perturbation with a channel-wise scale, and $\mathrm{M}_4$ further adapts this scale to spatial location and image content.
Both $\mathrm{M}_2$ and $\mathrm{M}_3$ match $\mathrm{M}_4$ in parameter count and forward FLOPs, providing matched baselines that control for gains from increased model size or computation.

In \cref{tab:ablation}, $\mathrm{M}_0$ removes GR, retains the direct synthesis path of the EF-LIC backbone, and shows performance close to EF-LIC.
Because $\mathrm{M}_0$ and $\mathrm{M}_1$--$\mathrm{M}_4$ are trained under the same conditions, $\mathrm{M}_0$ also provides a direct reference for comparing GR with the EF-LIC backbone.
$\mathrm{M}_4$ reduces FID by $4.2$ to $7.8$ relative to $\mathrm{M}_2$ and by $1.5$ to $4.9$ relative to $\mathrm{M}_3$, while achieving lower LPIPS at every evaluated rate.
Since $\mathrm{M}_2$ and $\mathrm{M}_3$ match $\mathrm{M}_4$ in parameter count and forward FLOPs, these results show that the complete GR configuration improves reconstruction performance on the lightweight EF-LIC backbone.
We select $\mathrm{M}_4$ for GR.

To further understand the role of the perturbation, we set only the sampled term $\bm\beta\odot\bm\epsilon$ to zero during $\mathrm{M}_4$ inference and observe only a small performance change.
Together with the gains over deterministic $\mathrm{M}_2$, this result supports an important role of adaptive perturbation as latent noise regularization during training.
The gains over $\mathrm{M}_3$ further show the benefit of adapting the perturbation scale to spatial location and image content.
\Cref{app:gr_ablation} specifies the configuration training protocol, and \cref{app:gr_stability} reports the inference perturbation results.

\subsection{Edge-Device Deployment}
\label{sec:edge}

We evaluate \ours on a 16 GB NVIDIA Jetson Orin NX.
Among the compared methods, \ours achieves the lowest decoding latency at all four resolutions and the lowest encoding latency from 1080p to 4K in \cref{tab:edge_devices}.
Measurements use MAXN SUPER mode, batch size one, and synchronized wall-clock timing.
\ours remains deployable at 4K, where OneDC and AEIC run out of memory and CoD-Lite decoding exceeds 10 seconds.
These results demonstrate effective low-latency deployment on edge hardware.

\section{Limitations}
\label{sec:limitations}
Our evaluation spans 0.0356 to 0.1426 BPP, with the lowest point below the 0.05 BPP ultra-low bitrate threshold~\citep{diffusion-StableCodec,aeic2026}.
Extending lightweight GIC to even lower bitrates requires greater decoder generative capacity as less fine detail is transmitted.
Future work can strengthen lightweight priors and synthesis capacity while retaining fast inference.
Our latency results characterize the complete deployed codec, including P-RVQ reorganization and CUDA acceleration.
We do not claim the fastest backbone-only FP32 inference under identical backends.
Comparisons retain official or released implementations and available acceleration paths, reflecting system-level performance.

\section{Conclusion}
We introduced \ours, an efficient GIC combining GR and reorganized P-RVQ for competitive reconstruction performance and fast inference.
GR refines quantized latents using entropy context, while adaptive perturbations provide latent noise regularization during training.
Reorganized P-RVQ replaces serial high dimensional residual updates with low dimensional lookups to reduce sequential computation.
Experiments across four datasets show consistent gains in LPIPS, DISTS, and FID over OneDC, together with fast inference on both an NVIDIA A100 and Jetson Orin NX.

{\small
\bibliographystyle{ieeenat_fullname}
\bibliography{references}
}

\clearpage
\appendix
\section*{Appendix}
In the appendix, we provide the following:
\begin{itemize}
  \item \Cref{app:rvq} derives the reorganized P-RVQ, analyzes its complexity, and reports implementation details and latency results.
  \item \Cref{app:details} describes the model architecture, bitstream packing, padding, and nominal rate.
  \item \Cref{app:protocol} defines the dataset, metric, runtime, baseline, and GR ablation evaluation protocols.
  \item \Cref{app:additional_results} reports additional efficiency and quality comparisons, high-resolution results, and GR inference stability.
  \item \Cref{app:more_visuals} provides additional visual comparisons on all four evaluation datasets.
\end{itemize}
\section{Analysis of Reorganized P-RVQ}
\label{app:rvq}

The main text presents the key reorganization of P-RVQ with projection biases omitted for clarity.
This appendix restores the complete affine transformations, establishes the equivalence between the original and reorganized forms, analyzes their computation and storage, and reports the corresponding runtime results.

\subsection{Complete P-RVQ Reorganization}

We use the notation of \cref{sec:rvq_optimization}.
Let $\bm z\in\mathbb R^D$ denote the input to a P-RVQ module, covering both the hyper latent and the main latent groups.
The following equations describe one input vector, while $L$ denotes the number of input vectors when counting operations.

At stage $m$, the affine projection $(\bm A_m,\bm a_m)$ maps the current residual from $\mathbb R^D$ to the search space $\mathbb R^d$, where
$\bm A_m\in\mathbb R^{d\times D}$ and $\bm a_m\in\mathbb R^d$.
The codebook
$\bm E_m=[\bm E_m[1],\ldots,\bm E_m[K]]\in\mathbb R^{d\times K}$
contains $K$ codewords.
The reconstruction projection $(\bm B_m,\bm b_m)$ maps the selected codeword back to $\mathbb R^D$, where
$\bm B_m\in\mathbb R^{D\times d}$ and $\bm b_m\in\mathbb R^D$.
The value of $K$ may differ among the five P-RVQ modules.

Starting from $\hat{\bm z}_0=\bm0$, let $\hat{\bm z}_m$ denote the partial reconstruction after stage $m$, and let $\hat{\bm z}=\hat{\bm z}_M$ denote the final reconstruction.

\paragraph{Original execution.}

The original P-RVQ projects the current residual for nearest neighbor search and immediately adds the selected reconstruction to the high dimensional state:

\begin{align}
  i_m
  &=
  \arg\min_{1\leq j\leq K}
  \left\|
    \bm A_m(\bm z-\hat{\bm z}_{m-1})
    +\bm a_m
    -\bm E_m[j]
  \right\|_2^2,
  \\
  \hat{\bm z}_m
  &=
  \hat{\bm z}_{m-1}
  +\bm B_m\bm E_m[i_m]
  +\bm b_m.
  \label{eq:appendix_original_rvq}
\end{align}

The final reconstruction is therefore

\begin{equation}
  \hat{\bm z}
  =
  \hat{\bm z}_M
  =
  \sum_{m=1}^{M}
  \left(
    \bm B_m\bm E_m[i_m]
    +\bm b_m
  \right).
  \label{eq:appendix_original_decode}
\end{equation}

Before stage $m+1$, the selected codeword must pass through $\bm B_m$ and update the state in $\mathbb R^D$.
The next stage then projects this updated residual back to $\mathbb R^d$.
This produces the
$\mathbb R^D\rightarrow\mathbb R^d\rightarrow\mathbb R^D$
dependency path shown in \cref{eq:projected_rvq_original}.

\paragraph{Reorganized execution.}

Expanding the partial reconstruction inside the search projection gives

\[
  \begin{aligned}
  &
  \bm A_m(\bm z-\hat{\bm z}_{m-1})
  +\bm a_m
  \\
  &\quad=
  \bm A_m\bm z
  +\bm a_m
  -
  \sum_{\ell<m}
  \bm A_m
  \left(
    \bm B_\ell\bm E_\ell[i_\ell]
    +\bm b_\ell
  \right).
  \end{aligned}
\]

All parameters and codewords in the terms from earlier stages are fixed after training.
We therefore precompute the $d$ dimensional lookup table

\begin{equation}
  \begin{aligned}
  \bm F_{\ell\rightarrow m}[j]
  &=
  \bm A_m
  \left(
    \bm B_\ell\bm E_\ell[j]
    +\bm b_\ell
  \right)
  \\
  &=
  (\bm A_m\bm B_\ell)\bm E_\ell[j]
  +\bm A_m\bm b_\ell,
  \qquad
  \ell<m.
  \end{aligned}
  \label{eq:appendix_rvq_cross_table}
\end{equation}

This restores the bias term omitted from the simplified expression in
\cref{eq:projected_rvq_rewrite}.
Using these tables, the reorganized recursion becomes

\begin{align}
  i_m
  &=
  \arg\min_{1\leq j\leq K}
  \Bigl\|
    \bm A_m\bm z
    +\bm a_m
    \nonumber\\
  &\qquad{}-
    \sum_{\ell<m}
    \bm F_{\ell\rightarrow m}[i_\ell]
    -
    \bm E_m[j]
  \Bigr\|_2^2,
  \label{eq:appendix_optimized_rvq}\\
  \hat{\bm z}_m
  &=
  \hat{\bm z}_{m-1}
  +\bm B_m\bm E_m[i_m]
  +\bm b_m.
  \nonumber
\end{align}

The source projections for all stages are evaluated together:

\begin{equation}
  \begin{bmatrix}
    \bm s_1^{(0)}\\[-2pt]
    \vdots\\[-2pt]
    \bm s_M^{(0)}
  \end{bmatrix}
  =
  \begin{bmatrix}
    \bm A_1\\[-2pt]
    \vdots\\[-2pt]
    \bm A_M
  \end{bmatrix}
  \bm z
  +
  \begin{bmatrix}
    \bm a_1\\[-2pt]
    \vdots\\[-2pt]
    \bm a_M
  \end{bmatrix}.
  \label{eq:appendix_rvq_stacked_projection}
\end{equation}

At stage $m$, the corresponding search state is

\begin{equation}
  \bm s_m
  =
  \bm s_m^{(0)}
  -
  \sum_{\ell<m}
  \bm F_{\ell\rightarrow m}[i_\ell]
  =
  \bm A_m(\bm z-\hat{\bm z}_{m-1})
  +\bm a_m.
  \label{eq:appendix_rvq_search_equivalence}
\end{equation}

After selecting $i_\ell$, the later search states are updated through
$\bm F_{\ell\rightarrow m}[i_\ell]$ for $m>\ell$.
Equation~\ref{eq:appendix_rvq_search_equivalence} shows that the original and reorganized forms present the same search vector to every stage in real arithmetic.
Starting from the first stage, the same nearest neighbor decision therefore produces the same update for all later stages.
Induction over $m$ preserves all greedy indices and the final reconstruction.

The reorganized execution does not need to materialize the high dimensional partial reconstruction between search stages.
It maintains the low dimensional states $\bm s_m$ until all indices are selected, and then gathers and sums
$\bm B_m\bm E_m[i_m]+\bm b_m$
to obtain \cref{eq:appendix_original_decode}.
The greedy stage order remains unchanged, while the high dimensional reconstruction is removed from the critical dependency path.

For nearest neighbor search, the part of the squared distance that depends on candidate $j$ is

\[
  \begin{aligned}
  i_m
  &=
  \arg\min_j
  \left\|
    \bm s_m-\bm E_m[j]
  \right\|_2^2
  \\
  &=
  \arg\min_j
  \left(
    \left\|
      \bm E_m[j]
    \right\|_2^2
    -
    2\bm s_m^{\mathsf T}\bm E_m[j]
  \right),
  \end{aligned}
\]

because $\|\bm s_m\|_2^2$ is constant across candidates.
The codeword norms and codebook transposes can therefore be stored before inference.

\subsection{Complexity and Parallelism}

The reorganization preserves the projection used for search and the nearest neighbor computation.
For $L$ input vectors, the shared multiply-accumulate counts are

\begin{align*}
  \mathcal C_{\mathrm{proj}}^{\mathrm{MAC}}
  &=
  L\sum_{m=1}^{M}Dd
  =
  LMDd,
  \\
  \mathcal C_{\mathrm{search}}^{\mathrm{MAC}}
  &=
  L\sum_{m=1}^{M}Kd
  =
  LMKd,
  \\
  \mathcal C_{\mathrm{common}}^{\mathrm{MAC}}
  &=
  LM(D+K)d.
\end{align*}

Both forms also require $LMd$ additions for projection biases and $LM(K-1)$ candidate comparisons.
In the evaluated codec, the four main latent P-RVQ modules use
$D=256$, $d=16$, and
$K\in\{1024,512,256,128\}$.
The hyper latent P-RVQ uses
$D=128$, $d=8$, and $K=1024$.
Thus, all five modules satisfy $D=16d$.

\paragraph{Reconstruction cost.}

The original execution evaluates
$\bm B_m\bm E_m[i_m]+\bm b_m$
online for every selected codeword.
The reorganized execution evaluates these projections for all codewords during checkpoint conversion and stores the resulting reconstruction tables.
For one P-RVQ module,

\begin{align*}
  \mathcal C_{\mathrm{out,online}}^{\mathrm{MAC}}
  &=
  LMDd,
  \\
  \mathcal C_{\mathrm{out,offline}}^{\mathrm{MAC}}
  &=
  MKDd,
  \\
  \mathcal S_{\mathrm{decode}}
  &=
  MKD.
\end{align*}

The original online path additionally performs $LMD$ bias additions.
The reorganized online path replaces the projections with table lookups and requires $L(M-1)D$ additions to sum the selected reconstruction vectors.
Both forms produce the same reconstruction in \cref{eq:appendix_original_decode}.

\paragraph{State update cost.}

The main reduction occurs in the sequential state updates.
The corresponding addition counts and their ratio are

\begin{align}
  \mathcal C_{\mathrm{high}}^{\mathrm{add}}
  &=
  L\sum_{m=2}^{M}D
  =
  L(M-1)D,
  \\
  \mathcal C_{\mathrm{low}}^{\mathrm{add}}
  &=
  L\sum_{m=2}^{M}(m-1)d
  =
  \frac{LM(M-1)d}{2},
  \\
  \frac{
    \mathcal C_{\mathrm{high}}^{\mathrm{add}}
  }{
    \mathcal C_{\mathrm{low}}^{\mathrm{add}}
  }
  &=
  \frac{2D}{Md}.
  \label{eq:appendix_state_costs}
\end{align}

For $M=4$ and $D=16d$, the original and reorganized state update counts become $48Ld$ and $6Ld$, respectively.
The reorganization therefore uses $8\times$ fewer additions for state propagation.
The projection MACs, nearest neighbor MACs, and number of ordered greedy decisions remain unchanged.
This reduction characterizes the sequential state update path rather than the complete P-RVQ runtime.

\paragraph{Parallelism.}

The original schedule exposes one source projection at each stage.
The reorganized schedule evaluates the $M$ source projections together.
After selecting $i_\ell$, the lookup updates for all later stages can also be applied concurrently.
The reorganized execution therefore exposes all $M$ source projections and up to $M-1$ parallel low dimensional updates, while retaining the same $M$ ordered nearest neighbor decisions.

\paragraph{Lookup storage.}

The cross stage tables are constructed once after training.
Their offline computation and nonzero storage for one P-RVQ module are

\begin{align*}
  \mathcal C_{\mathrm{cross,offline}}^{\mathrm{MAC}}
  &=
  KDd
  \sum_{m=2}^{M}(m-1)
  =
  \frac{M(M-1)KDd}{2},
  \\
  \mathcal S_{\mathrm{cross,nonzero}}
  &=
  Kd
  \sum_{m=2}^{M}(m-1)
  =
  \frac{M(M-1)Kd}{2}.
\end{align*}

For $M=4$, six pairs satisfy $\ell<m$.
Across all five P-RVQ modules, the corresponding nonzero content is

\[
  \begin{aligned}
  \mathcal S_{\mathrm{nonzero}}^{\mathrm{all}}
  &=
  6
  \bigl(
    1024\!\cdot\!16
    +
    512\!\cdot\!16
    +
    256\!\cdot\!16
    \\
  &\qquad{}+
    128\!\cdot\!16
    +
    1024\!\cdot\!8
  \bigr)
  \\
  &=
  233{,}472.
  \end{aligned}
\]

The implementation uses a dense table with $M^2=16$ stage slots, giving

\[
  \begin{aligned}
  \mathcal S_{\mathrm{dense}}^{\mathrm{all}}
  &=
  16
  \bigl(
    1024\!\cdot\!16
    +
    512\!\cdot\!16
    +
    256\!\cdot\!16
    \\
  &\qquad{}+
    128\!\cdot\!16
    +
    1024\!\cdot\!8
  \bigr)
  \\
  &=
  622{,}592.
  \end{aligned}
\]

The dense table has shape $(M,K,Md)$ and occupies 1.19 MiB in FP16 or 2.38 MiB in FP32 across the five P-RVQ modules.
This storage is prepared before inference and contains the static lookup values used by the reorganized search.

\subsection{Implementation and Runtime Evaluation}

\paragraph{CUDA implementation.}

Checkpoint conversion materializes the stacked search projection, codeword norms, codebook transposes, reconstruction tables, and the cross stage lookup tables in \cref{eq:appendix_rvq_cross_table}.
At runtime, the source projections in \cref{eq:appendix_rvq_stacked_projection} are evaluated with one matrix multiplication.

A fused CUDA kernel then processes the ordered P-RVQ stages.
Candidate dot products, distance evaluation, parallel reduction for index selection, table lookup, and updates to the later low dimensional states are performed within one kernel launch.
The kernel preserves the ordered greedy decisions while parallelizing candidate evaluation and the updates to future search states.
This avoids separate search launches at every stage and avoids materializing complete distance and update tensors.

When reconstruction is required, a second kernel gathers the selected high dimensional vectors, sums them, and writes the reconstructed output.
During decompression, these reconstruction operations are further combined with the surrounding codec operations.
This codec integration lies outside the isolated P-RVQ timing reported below.

\paragraph{Benchmark protocol.}

We evaluate the original Torch FP32 implementation, the reorganized Torch implementations in FP32 and FP16, and the CUDA FP16 implementation on one NVIDIA A100 with batch size one.
All implementations use weights converted from the same checkpoints.

We evaluate $M\in\{1,2,3,4\}$ at latent sizes corresponding to $512\times768$, 1080p, 2K, and 4K images, giving 16 settings in total.
Each latency is the median of 100 synchronized repetitions after 20 warmup runs.

The reported latency isolates P-RVQ execution.
Complete codec latency is reported in \cref{tab:main_results,tab:edge_devices,tab:highres}.

\paragraph{Runtime results.}

Under the same Torch FP32 backend and precision, the reorganized P-RVQ achieves a $3.14\times$ average speedup over the original implementation across the 16 evaluated settings.
The reorganized Torch FP16 implementation achieves a $2.97\times$ average speedup.
These matched Torch results isolate the effect of the algebraic reorganization and numerical precision.

The CUDA FP16 implementation further increases the average speedup to $7.15\times$ relative to the original FP32 implementation.
This number includes the algebraic reorganization, FP16 execution, and fused CUDA kernels.
\Cref{tab:rvq_latency} reports the latency and speedup for every resolution and P-RVQ depth.

At the highest evaluated Kodak rate, the P-RVQ latency during compression is
11.2824 ms for the original FP32 implementation,
3.2435 ms for the reorganized Torch FP32 implementation,
and 0.6001 ms for the CUDA FP16 implementation.
The matched FP32 reorganization gives a $3.48\times$ speedup at this operating point.
The CUDA FP16 implementation gives an $18.80\times$ speedup.

For complete encoder attribution, we construct an original P-RVQ reference by restoring the latency difference between the original and reorganized FP32 P-RVQ to the measured Torch FP32 encoder.
Using the reduction from this reference to the final CUDA FP16 encoder as a common denominator, the algebraic reorganization accounts for $23.54\%$ of the total encoder latency reduction.
The complete P-RVQ acceleration, including the CUDA implementation, accounts for $31.28\%$.

\begin{table*}[t]
\centering
\caption{P-RVQ latency on an NVIDIA A100.
Each entry reports the median CUDA event latency in ms followed by speedup over the corresponding original FP32 result.
All implementations use weights converted from the same checkpoints.}
\label{tab:rvq_latency}
\begin{tabular}{ccrrrr}
\toprule
Resolution & $M$ & Original FP32 & Reorganized FP32 & Reorganized FP16 & CUDA FP16 \\
\midrule
\multirow{4}{*}{$512\!\times\!768$} & 1 & $3.84 / 1.00\times$ & $1.54 / 2.50\times$ & $1.54 / 2.49\times$ & $0.93 / 4.12\times$ \\
 & 2 & $5.38 / 1.00\times$ & $1.74 / 3.09\times$ & $1.74 / 3.09\times$ & $0.64 / 8.43\times$ \\
 & 3 & $10.80 / 1.00\times$ & $3.29 / 3.29\times$ & $3.33 / 3.24\times$ & $0.87 / 12.38\times$ \\
 & 4 & $14.09 / 1.00\times$ & $4.03 / 3.49\times$ & $4.15 / 3.40\times$ & $0.87 / 16.12\times$ \\
\midrule
\multirow{4}{*}{1080p} & 1 & $2.82 / 1.00\times$ & $1.14 / 2.47\times$ & $1.14 / 2.48\times$ & $0.67 / 4.18\times$ \\
 & 2 & $5.42 / 1.00\times$ & $1.77 / 3.07\times$ & $1.83 / 2.97\times$ & $0.66 / 8.26\times$ \\
 & 3 & $8.37 / 1.00\times$ & $2.43 / 3.45\times$ & $2.56 / 3.27\times$ & $0.78 / 10.67\times$ \\
 & 4 & $10.59 / 1.00\times$ & $3.03 / 3.49\times$ & $3.20 / 3.31\times$ & $0.95 / 11.16\times$ \\
\midrule
\multirow{4}{*}{2K} & 1 & $2.88 / 1.00\times$ & $1.12 / 2.58\times$ & $1.16 / 2.47\times$ & $0.68 / 4.25\times$ \\
 & 2 & $5.40 / 1.00\times$ & $1.74 / 3.09\times$ & $1.96 / 2.76\times$ & $0.87 / 6.23\times$ \\
 & 3 & $8.34 / 1.00\times$ & $2.51 / 3.33\times$ & $2.55 / 3.27\times$ & $1.18 / 7.10\times$ \\
 & 4 & $10.83 / 1.00\times$ & $3.06 / 3.55\times$ & $3.27 / 3.31\times$ & $1.48 / 7.32\times$ \\
\midrule
\multirow{4}{*}{4K} & 1 & $2.93 / 1.00\times$ & $1.13 / 2.61\times$ & $1.16 / 2.54\times$ & $0.98 / 3.00\times$ \\
 & 2 & $5.97 / 1.00\times$ & $1.92 / 3.11\times$ & $3.00 / 1.99\times$ & $1.67 / 3.58\times$ \\
 & 3 & $8.61 / 1.00\times$ & $2.45 / 3.51\times$ & $2.49 / 3.46\times$ & $2.30 / 3.74\times$ \\
 & 4 & $11.42 / 1.00\times$ & $3.09 / 3.69\times$ & $3.24 / 3.52\times$ & $2.99 / 3.82\times$ \\
\midrule
\multicolumn{2}{c}{Mean speedup} & $1.00\times$ & $3.14\times$ & $2.97\times$ & $\mathbf{7.15\times}$ \\
\bottomrule
\end{tabular}
\end{table*}

\paragraph{Numerical consistency.}

The original and reorganized Torch FP32 implementations select identical indices, with a maximum absolute reconstruction difference of $4.77\times10^{-7}$.
Under FP16, the reorganized Torch and CUDA implementations differ on only 0.2293\% of the evaluated indices.
These differences mainly occur between nearby codewords and have little effect on the final reconstruction.

\section{Architecture and Bitstream Details}
\label{app:details}

\subsection{Backbone and Context Pathway}

\ours follows the lightweight backbone and context pathway of EF-LIC~\citep{efic2026}.
\Cref{tab:arch} gives the complete implemented topology.

For clarity, \cref{sec:overview} absorbs the normalization used in the implementation into the analysis and synthesis transforms.
The encoder first computes the latent before normalization,
\[
    \bm v=g_a(\bm x),
    \qquad
    \bm z=h_a(\bm v).
\]
After quantizing the hyper latent, the hyper decoder produces the normalization scale and context feature,
\[
    (\bm q_{\mathrm{enc}},\bm\phi)=h_s(\hat{\bm z}),
    \qquad
    \bm y=\bm v\oslash\bm q_{\mathrm{enc}}.
\]
The normalized latent $\bm y$ is then processed by the groupwise affine projections and P-RVQ.

At the decoder, the same hyper decoder gives
\[
    (\bm q_{\mathrm{dec}},\bm\phi)=h_s(\hat{\bm z}).
\]
For a fixed checkpoint and bitstream, $\bm q_{\mathrm{enc}}=\bm q_{\mathrm{dec}}$, and reconstruction is obtained as
\[
    \hat{\bm x}
    =
    g_s(\bm q_{\mathrm{dec}}\odot\tilde{\bm y}).
\]
The notation in \cref{sec:overview} absorbs these deterministic scales into $g_a$ and $g_s$.
The normalization scale $\bm q_{\mathrm{dec}}$ and entropy context scale $\bm\sigma$ are lower bounded by $0.11$.

The main latent uses the same four group quadtree context model as EF-LIC and DCVC-RT~\citep{efic2026,dcvc-rt}.
Each group contains one quarter of the spatial positions and is recovered using the previously decoded groups.
The GR head receives the recovered latent and entropy context parameters $(\hat{\bm y},\bm\mu,\bm\sigma)$ and predicts the offset $\bm\delta$ and perturbation scale $\bm\beta$.
Each P-RVQ depth $M\in\{1,2,3,4\}$ defines one separately trained operating point with its corresponding GR head.

\begin{table*}[t]
\centering
\caption{Detailed \ours architecture.
Conv, DCB, and SP denote convolution, DepthConvBlock, and subpixel upsampling, respectively.
The analysis and synthesis depths count the repeated body DCBs, while the entry and final refinement blocks are listed separately.}
\label{tab:arch}
\begin{tabularx}{\textwidth}{
    >{\raggedright\arraybackslash}p{0.17\textwidth}
    >{\raggedright\arraybackslash}X
    >{\raggedright\arraybackslash}p{0.26\textwidth}}
\toprule
Module & Implemented topology & Output or role \\
\midrule

Analysis transform
&
$8\times$ pixel unshuffle,
$192\!\to\!368$ entry DCB,
six 368-channel body DCBs,
$3\!\times\!3$ Conv with stride 2 to 256 channels
&
Main latent at $1/16$ resolution
\\

Hyper analysis
&
$1\!\times\!1$ Conv $256\!\to\!512$,
two 512-channel DCBs,
two residual downsampling stages with stride 2,
final 128-channel projection
&
Hyper latent at $1/64$ resolution
\\

Hyper synthesis
&
SP $128\!\to\!512$,
one DCB,
SP $512\!\to\!512$,
two DCBs,
$1\!\times\!1$ Conv to 512 channels
&
256-channel normalization scale $\bm q_{\mathrm{dec}}$
and 256-channel context feature $\bm\phi$
\\

Context path
&
Quadtree support projection,
projection of previously decoded groups,
one group adaptor DCB,
three shared 512-channel DCBs,
final $1\!\times\!1$ Conv
&
256-channel entropy context mean $\bm\mu$
and scale $\bm\sigma$
\\

GR head
&
$1\!\times\!1$ Conv $768\!\to\!768$,
three 768-channel DCBs,
$1\!\times\!1$ Conv $768\!\to\!512$
&
256-channel offset $\bm\delta$
and adaptive perturbation scale $\bm\beta$
\\

Synthesis transform
&
SP $256\!\to\!368$,
twelve 368-channel body DCBs,
one final refinement DCB,
$1\!\times\!1$ Conv $368\!\to\!192$,
$8\times$ pixel shuffle
&
RGB reconstruction
\\

DCB
&
Pointwise mixing,
weighted sigmoid linear unit (SiLU),
$3\!\times\!3$ depthwise Conv,
pointwise projection,
fourfold feedforward expansion and projection
&
Residual feature transformation
\\

\bottomrule
\end{tabularx}
\end{table*}

\subsection{Fixed-Width Index Packing}
\label{app:packing}

Each power of two codebook in \cref{tab:quantizers} uses a fixed index width according to the size of its quantization alphabet~\citep{q}.
The hyper latent uses 10 bits per index, while the four main latent groups use 10, 9, 8, and 7 bits.

\begin{table}[t]
\centering
\caption{P-RVQ configuration.
The four main latent quantizers operate on quadtree groups.
The fifth quantizes the hyper latent.
``Bits per stage'' is the fixed index width.}
\label{tab:quantizers}
\begin{tabular}{lrrrrr}
\toprule
Stream & $y_1$ & $y_2$ & $y_3$ & $y_4$ & $z$ \\
\midrule
Input dim.
& 256 & 256 & 256 & 256 & 128 \\
Code dim.
 & 16  & 16  & 16  & 16  & 8 \\
Entries    & 1024& 512 & 256 & 128 & 1024 \\
Bits per stage & 10  & 9   & 8   & 7   & 10 \\
\bottomrule
\end{tabular}
\end{table}

The multistage structure and low dimensional search spaces follow the P-RVQ configuration of EF-LIC~\citep{efic2026}.
Stored indices are flattened in $(\text{stage},\text{batch},\text{height},\text{width})$ order and written with the least significant bit first.
The five streams are ordered as
$Q_{\bm z}\rightarrow Q_1\rightarrow Q_2\rightarrow Q_3\rightarrow Q_4$.

A standalone bitstream stores the original image height and width as two unsigned 16 bit integers, followed by the five index streams.
No additional side information is transmitted.
We report payload BPP and exclude this 4 byte image size header.
For a $512\!\times\!768$ image, the header contributes only

\[
    R_{\mathrm{meta}}
    =
    \frac{32}{512\cdot768}
    =
    8.14\times10^{-5}\ \mathrm{bpp}.
\]

\subsection{Padding and Cropping}
\label{app:padding}

The encoder pads the input on the right and bottom so that both spatial dimensions are multiples of 32.
The hyper analysis path additionally pads the main latent to a multiple of four, while the main latent path retains its original logical shape.
The quadtree groups, main latent P-RVQ modules, GR head, and synthesis transform therefore operate on this original latent shape.
The decoder removes the corresponding padding and returns the reconstruction at the original image size.

\subsection{Nominal Rate}
\label{app:nominal_rate}

The fixed index widths give an exact nominal payload rate before boundary padding.
At each P-RVQ stage, the four main latent groups contribute

\[
    R_y
    =
    \frac{10+9+8+7}{4\cdot16^2}
    =
    0.033203125\ \mathrm{bpp},
\]

while the hyper latent contributes

\[
    R_z
    =
    \frac{10}{64^2}
    =
    0.00244140625\ \mathrm{bpp}.
\]

Thus, with $M$ active P-RVQ stages,

\begin{equation}
    R_M
    =
    M(R_y+R_z)
    =
    0.03564453125M\ \mathrm{bpp}.
  \label{eq:nominal_rate}
\end{equation}

Boundary padding changes the number of transmitted indices and causes the small deviations from these nominal rates.

\section{Evaluation Protocol}
\label{app:protocol}

\subsection{Datasets and Objective Evaluation}

Kodak, Tecnick, DIV2K validation, and CLIC 2020 Professional test contain 24, 100, 100, and 250 images, respectively.
All reconstructed images are evaluated at their original image resolution.

\paragraph{Objective evaluation.}

We use standardized objective metrics as the primary quantitative evaluation.
Subjective user studies can provide complementary perceptual evidence, but their results depend on the participant population, viewing conditions, presentation protocol, and selected test images, and naturally exhibit observer variability.
These factors make subjective scores less suitable for reproducible comparisons across many methods, datasets, and operating points.
Standardized objective metrics instead provide repeatable measurements under a fixed evaluation protocol.

This practice is also common in generative image compression.
Representative methods including MS-ILLM, GLC, Control-GIC, DLF, OneDC, OSCAR, and RDVQ primarily evaluate reconstruction using objective metrics such as LPIPS, DISTS, and FID without reporting a subjective user study~\citep{ms-illm,glc2024,cgic,vq-DLF,diffusion-xue2025one,oscar,rdvq2026}.
We therefore use LPIPS and DISTS to measure reference-based perceptual similarity, PSNR to measure pixel fidelity, and FID to measure distribution-level alignment, together with qualitative reconstructions for visual comparison.

LPIPS is computed with pretrained VGG features after mapping RGB inputs to $[-1,1]$.
DISTS receives RGB inputs in $[0,1]$, and PSNR is computed in RGB.

\paragraph{FID.}
\label{app:pfid}

Following the patch evaluation protocol used for low-rate generative image compression~\citep{glc2024}, we compute FID from local image patches.
Reference and reconstructed RGB images in $[0,1]$ are rounded to unsigned 8-bit values and passed to the TorchMetrics implementation of FID~\citep{detlefsen2022torchmetrics,fid}.

Kodak uses $64\times64$ patches, while Tecnick, DIV2K, and CLIC 2020 use $256\times256$ patches.
We extract one non-overlapping patch grid and a second grid shifted by half a patch along both spatial axes.
This increases the number of Inception samples and improves the stability of distribution-level evaluation on the 24-image Kodak set.
All compared methods use the same FID pipeline.

\paragraph{BD-rate.}

We use BD-rate~\citep{bd-rate} to summarize the average rate difference between two rate-quality curves at matched reconstruction quality.
For each metric, operating points are ordered by the metric value, and pairs without a valid common interval are left unreported.

Let $q$ denote the metric value.
We use monotone PCHIP interpolation~\citep{fritsch1984pchip} for the natural logarithm of bitrate.
Let $f_{\mathrm A}(q)$ and $f_{\mathrm T}(q)$ denote the interpolated curves of the anchor and tested codec over the common interval $[q_{\min},q_{\max}]$.
The reported BD-rate is

\begin{equation}
  \begin{aligned}
  \mathrm{BD\text{-}rate}
  &=
  100
  \Biggl[
    \exp\!\Biggl(
      \frac{1}{q_{\max}-q_{\min}}
      \\
  &\quad
      \int_{q_{\min}}^{q_{\max}}
      \left(
        f_{\mathrm T}(q)-f_{\mathrm A}(q)
      \right)
      \,\mathrm dq
    \Biggr)
    -1
  \Biggr].
  \end{aligned}
  \label{eq:bdrate_protocol}
\end{equation}

OneDC is the anchor for the main comparisons.
Negative values indicate bitrate savings over the common interval.
The numerical tables report pairwise BD-rate, while the rate-quality curves show the corresponding behavior over the evaluated operating points.

\subsection{Runtime and Model Cost Protocol}
\label{app:runtime}

All A100 codec runtimes use batch size one on one NVIDIA A100 GPU.
For \ours, CPU payload processing is performed on an AMD EPYC 7763.

The latency comparison retains the numerical precision, inference backend, and acceleration path provided by each available implementation.
The reported values therefore characterize realized system latency rather than architecture-only speed under an identical backend.
EF-LIC and ProGIC follow their published FP32 evaluation settings~\citep{efic2026,progic2026}, while the remaining methods retain their released inference settings.

The highly optimized implementation of \ours combines reorganized P-RVQ with fused CUDA operators following DCVC-UF~\citep{dcvc-uf}.
It uses FP16 execution and CUDA Graphs together with custom kernels for quantization and reconstruction.
The remaining convolution operations use optimized ATen and cuDNN kernels.

The reported encoding and decoding times cover the complete codec path.
Encoding includes index generation, transfer to the CPU, and bitstream packing.
Decoding includes bitstream unpacking, transfer to the GPU, index recovery, one GR realization, and image synthesis.
Timing starts after model loading, input construction, and CUDA Graph capture.
We use two warmup runs followed by ten synchronized repetitions.

For \ours, the reported latency and model cost are averaged over four independently trained operating points at
$0.0356$, $0.0713$, $0.1069$, and $0.1426$ BPP.
The same averaging convention is used for the A100 and Jetson results.

\paragraph{Parameter accounting across operating points.}

We report the parameter count of the complete codec deployed at one evaluated operating point.
Each of the four independently trained \ours checkpoints contains 79.389M parameters.
The reported 79.389M therefore represents one deployed model rather than the combined storage of all four checkpoints.

This convention is also applied to methods that use separate checkpoints for different target rates.
Among the compared methods, ResULIC, StableCodec, RDEIC, OneDC, MS-ILLM, DLF, DiT-IC, AEIC, and CoD-Lite use separately trained or fine-tuned model weights for different operating points~\citep{resulic2025,diffusion-StableCodec,rdeic,diffusion-xue2025one,ms-illm,vq-DLF,ditic2026,aeic2026,codlite2026}.

We use the same parameter accounting rule for both types of codec.
The tables report the complete model required for the evaluated operating point and do not sum multiple checkpoints across the rate curve.
The storage required to retain all supported operating points is a separate deployment property from the parameter count of one deployed codec.

FLOPs are measured for one encoder and decoder pass with batch size one on the corresponding PyTorch FP32 path under the profiler-supported operator scope.
We count one multiply-accumulate operation as two FLOPs.
The same pass definition and arithmetic convention are used for all methods.
Peak memory denotes the maximum allocated GPU memory including the loaded model and active codec buffers.

\subsection{Baseline Evaluation Protocol}
\label{app:baseline_protocol}

Except where noted, we evaluate released checkpoints on the same NVIDIA A100 with batch size one.
Each method follows the coding and inference procedure provided by its official implementation.

EF-LIC and ProGIC use their published FP32 settings~\citep{efic2026,progic2026}.
OneDC follows its official path with BF16 autocast for the feedforward and codec networks while retaining the large VAE in FP32~\citep{diffusion-xue2025one}.
CoD-Lite uses its official $0.0312$ BPP checkpoint with BF16 inference~\citep{codlite2026}.
Under our full codec profiling scope, this checkpoint gives 771.69G FLOPs at $512\times768$.
The CoD-Lite paper reports 535 kMAC per pixel, corresponding to 420.74G FLOPs under our arithmetic convention.
We use our common profiling scope for the system comparison and retain the published value as a separate reference.
Control-GIC uses its official granularity control, which we enumerate at intervals of 0.01, together with official tiling on DIV2K and CLIC 2020~\citep{cgic}.

For GLC~\citep{glc2024}, the released interface provides analysis and synthesis without a complete bitstream path.
For the high-resolution codec comparison, we additionally use a CompressAI 1.2.6 integration~\citep{compressai} that produces and decodes the actual bitstream.

For OSCAR, rate and quality evaluation uses CPU offloading on high-resolution images when required, while the resource tables report the direct execution result~\citep{oscar}.

CADC~\citep{cadc2026} and PICO~\citep{pico2026} do not provide the corresponding executable models used in our comparison.
We therefore use their published curves or released numerical results only when the metric definition and evaluation range are compatible with ours.

For CoD-Lite, measurements from the released implementation differ from the latency values reported in its paper.
At the $1024\times1920$ reference resolution, we measure 39.522 ms for encoding and 56.863 ms for decoding, compared with the reported 16 and 24 ms.
We therefore use measurements from the released code under our common hardware and timing protocol rather than mixing measured and separately reported runtime values.
The system comparisons use the official $0.0312$ BPP checkpoint specified by the published speed evaluation.

\subsection{GR Ablation Protocol}
\label{app:gr_ablation}

The configurations in \cref{tab:ablation} are trained independently from scratch at each bitrate.
All configurations use the same training data, loss terms and weights, AdamW settings, learning-rate schedule, and update budget.
Each ablation model is trained for 1M iterations.
The main operating-point results use separately trained models with the complete 2M iteration schedule.

$\mathrm{M}_0$ bypasses GR and directly synthesizes from the recovered quantized latent.
It retains the direct synthesis path of the EF-LIC backbone and provides a reference under the same training conditions used for the GR configurations.

$\mathrm{M}_1$ predicts an offset from the recovered latent.
$\mathrm{M}_2$ additionally uses entropy context for deterministic refinement.
$\mathrm{M}_3$ introduces Gaussian perturbation with one learned scale for each channel, shared across spatial locations.
$\mathrm{M}_4$ further adapts the perturbation scale to spatial location and image content.

Each crop for $\mathrm{M}_3$ and $\mathrm{M}_4$ uses one standard Gaussian draw to estimate the expected image objective.
$\mathrm{M}_2$ and $\mathrm{M}_3$ have the same reported parameter count and forward FLOPs as $\mathrm{M}_4$.
They provide matched references for evaluating stochastic perturbation and its adaptation to image content and spatial location.

\subsection{Stochastic Evaluation Protocol}
\label{app:gr_controls}

The main results use a fixed decoder seed of 1926 and reconstruct each bitstream once.
Stochastic sampling is applied only after deterministic RVQ index recovery, so the transmitted indices and payload do not depend on the decoder seed.

\paragraph{Inference perturbation ablation.}

To examine the role of the sampled perturbation after training, we use the same trained $\mathrm{M}_4$ checkpoints as in \cref{tab:ablation}.
We retain the recovered indices, learned offset, and all model parameters, and set only
$\bm\beta\odot\bm\epsilon$
in \cref{eq:gr_sample} to zero during inference.
The decoder therefore synthesizes from the learned refinement center

\[
  \tilde{\bm y}
  =
  \hat{\bm y}
  +
  \bm\delta.
\]

The corresponding results are reported in the additional experiments.

\paragraph{Decoder seed stability.}

To measure sensitivity to the sampled Gaussian perturbation, we keep the checkpoint, transmitted indices, input images, and all deterministic computations fixed and vary only the decoder seed.
We evaluate ten seeds on all Kodak images at each operating point.
For every seed, metrics are first averaged over the dataset before variation across seeds is measured.

The resulting stability measurements are reported together with the additional GR experiments.

\section{Additional Experimental Results}
\label{app:additional_results}

This section reports additional experimental results under the evaluation protocols defined in \cref{app:protocol}.
We first examine the efficiency of different \ours operating points and inference backends, then provide additional comparisons with existing codecs, and finally analyze the inference behavior of GR.

\subsection{Efficiency across Operating Points}

\begin{table*}[t]
\centering
\caption{Per operating point \ours complexity, CUDA latency, and peak memory on one A100.
FLOPs are reported in G for one encoder and decoder pass under the profiler-supported operator scope.
Latency is in ms and peak GPU memory is in GiB.}
\label{tab:gr_cb_complexity}
\begin{tabular}{llrrrrr}
\toprule
Resolution & BPP & Params & FLOPs & Enc.
& Dec.
& Mem.
\\
\midrule
\multirow{4}{*}{$512\!\times\!768$}
 & 0.0356 & 79.389 & 456.491 & 4.475 & 5.949 & 0.183 \\
 & 0.0713 & 79.389 & 456.531 & 4.523 & 5.947 & 0.185 \\
 & 0.1069 & 79.389 & 456.571 & 4.717 & 6.028 & 0.187 \\
 & 0.1426 & 79.389 & 456.610 & 4.674 & 6.009 & 0.189 \\
\midrule
\multirow{4}{*}{1080p}
 & 0.0356 & 79.389 & 2425.111 & 13.817 & 21.428 & 0.250 \\
 & 0.0713 & 79.389 & 2425.321 & 14.111 & 21.483 & 0.250 \\
 & 0.1069 & 79.389 & 2425.532 & 14.137 & 21.389 & 0.252 \\
 & 0.1426 & 79.389 & 2425.742 & 14.323 & 21.466 & 0.255 \\
\midrule
\multirow{4}{*}{2K}
 & 0.0356 & 79.389 & 4293.353 & 22.447 & 35.987 & 0.312 \\
 & 0.0713 & 79.389 & 4293.725 & 22.501 & 36.022 & 0.313 \\
 & 0.1069 & 79.389 & 4294.097 & 22.880 & 35.897 & 0.315 \\
 & 0.1426 & 79.389 & 4294.469 & 23.209 & 35.905 & 0.318 \\
\bottomrule
\end{tabular}
\end{table*}

\Cref{tab:gr_cb_complexity} reports the complexity and runtime of the four independently trained operating points at $512\times768$, 1080p, and 2K.

At $512\times768$, encoding latency ranges from 4.475 to 4.717 ms, while decoding latency ranges from 5.947 to 6.028 ms.
At 1080p, the encoding and decoding ranges span at most 0.506 and 0.094 ms, respectively.
At 2K, the corresponding ranges are 0.762 and 0.125 ms.

These results show that increasing the active P-RVQ depth changes runtime only modestly relative to the shared backbone.

\subsection{GR-GIC Backend Comparison}
\label{app:backends}

We further compare three implementations of the same \ours checkpoints to separate the effects of numerical precision and backend optimization.

\begin{table*}[t]
\centering
\caption{\ours backend comparison on one A100.
Latency is in ms and peak GPU memory is in GiB.
Online E2E denotes one consecutive pair of in-memory encode and decode calls.}
\label{tab:gr_backends}
\begin{tabular}{llrrrr}
\toprule
Resolution & Backend & Enc.
& Dec.
& Online E2E & Mem.
\\
\midrule
\multirow{3}{*}{$512\!\times\!768$}
 & PyTorch FP32 & 30.453 & 44.708 & 66.519 & 0.438 \\
 & PyTorch FP16 & 15.439 & 20.298 & 37.784 & 0.222 \\
 & CUDA FP16 & 4.597 & 5.983 & 10.353 & 0.186 \\
\midrule
\multirow{3}{*}{1080p}
 & PyTorch FP32 & 109.967 & 176.403 & 282.883 & 0.965 \\
 & PyTorch FP16 & 32.895 & 57.600 & 86.817 & 0.465 \\
 & CUDA FP16 & 14.097 & 21.442 & 35.567 & 0.252 \\
\midrule
\multirow{3}{*}{2K}
 & PyTorch FP32 & 185.468 & 312.454 & 495.019 & 1.446 \\
 & PyTorch FP16 & 55.968 & 95.262 & 145.070 & 0.691 \\
 & CUDA FP16 & 22.759 & 35.952 & 59.118 & 0.314 \\
\midrule
\multirow{3}{*}{4K}
 & PyTorch FP32 & 407.490 & 709.180 & 1114.092 & 2.909 \\
 & PyTorch FP16 & 120.631 & 221.758 & 327.342 & 1.373 \\
 & CUDA FP16 & 49.787 & 81.143 & 132.971 & 0.492 \\
\bottomrule
\end{tabular}
\end{table*}

At $512\times768$, the eager PyTorch FP32 implementation encodes in 30.453 ms and decodes in 44.708 ms.
Relative to the official OneDC implementation, these times are $2.53\times$ and $4.57\times$ lower, respectively.
The eager PyTorch FP32 implementation is also faster in both encoding and decoding than six of the eight displayed diffusion codecs, and this behavior remains consistent across the evaluated resolutions.

The CUDA FP16 implementation further reduces runtime relative to the PyTorch FP32 implementation.
Across the evaluated resolutions, encoding is accelerated by $6.62\times$ to $8.18\times$, while decoding is accelerated by $7.47\times$ to $8.74\times$.

The eager FP32 results show that low latency is already present before the additional CUDA acceleration, while the optimized backend provides a further reduction in runtime.

\subsection{Additional Baseline Comparisons}
\label{app:additional_baselines}

CADC and PICO do not provide the corresponding executable models used in our comparison.
Following the protocol in \cref{app:protocol}, we therefore compare against their published curves or released numerical results where the metric definitions and rate ranges are compatible.

\paragraph{CADC.}

At the shared lowest Kodak rate of 0.03564 BPP, interpolation of the published CADC curves gives approximately 23.15 dB PSNR and 0.087 DISTS~\citep{cadc2026}.
At the same rate, \ours achieves 23.35 dB PSNR and 0.075 DISTS.
This corresponds to a PSNR increase of approximately 0.20 dB and a DISTS reduction of approximately 0.012.

CADC reports LPIPS using AlexNet features, while our main evaluation uses VGG features.
We therefore restrict this comparison to the aligned PSNR and DISTS metrics.

\paragraph{PICO.}

PICO reports VGG-based LPIPS on CLIC 2020~\citep{pico2026}.
Over the common quality interval, \ours obtains a PICO-anchored LPIPS BD-rate of $-54.01\%$.
This corresponds to a 54.01\% average bitrate reduction at matched LPIPS over the evaluated common interval.

\subsection{Comparison with Lightweight Codecs}
\label{app:lightweight_comparison}

\begin{figure*}[t]
  \centering
  \includegraphics[width=0.80\textwidth]{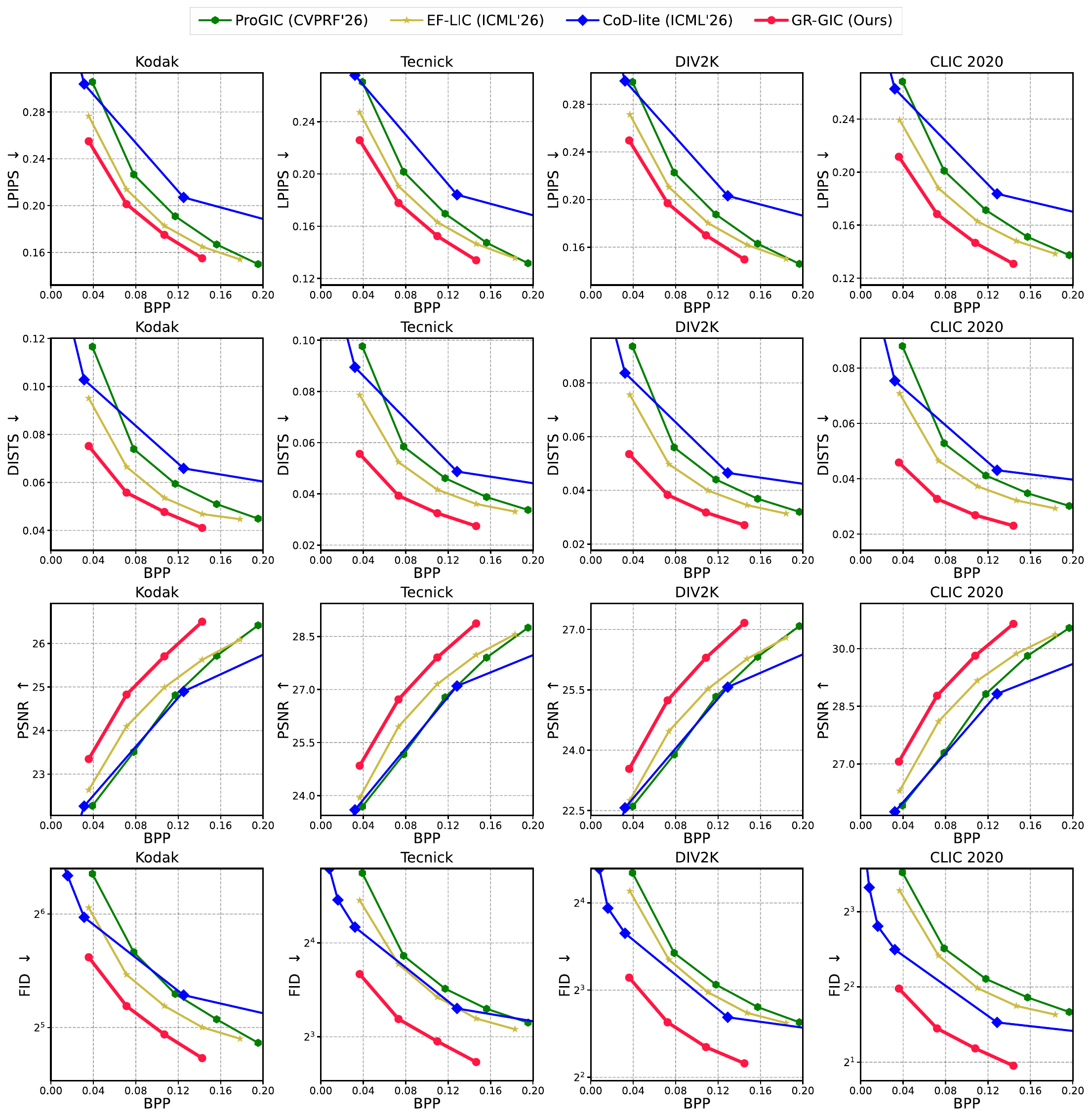}
  \caption{
  Rate and reconstruction quality of \ours and lightweight codecs on four datasets.
  \ours uses four operating points, while EF-LIC uses five.
  Lower LPIPS, DISTS, and FID and higher PSNR indicate better reconstruction.
  FID uses a base-2 logarithmic vertical axis.}
  \label{fig:lightweight_comparison}
\end{figure*}

\Cref{fig:lightweight_comparison} compares \ours with ProGIC, EF-LIC, and CoD-Lite across all four datasets and metrics.
At comparable rates, \ours provides the strongest displayed LPIPS, DISTS, PSNR, and FID curves among these lightweight codecs.

The four \ours operating points span approximately 0.036 to 0.143 BPP.
The highest-rate \ours point at 0.143 BPP uses a lower bitrate than the fifth EF-LIC point while achieving higher PSNR and lower DISTS and FID on every dataset.
Its LPIPS is comparable or lower.
The four \ours operating points therefore cover the shared comparison range without requiring an additional higher-rate point.

\subsection{Additional Rate and Quality Results}
\label{app:expanded_results}

\begin{table*}[t]
\centering
\caption{Pairwise OneDC-anchored BD-rate (\%).
Bold and underline mark the two lowest displayed values per dataset.}
\label{tab:bd_selected_appendix}
\begin{tabular}{lllrrrr}
\toprule
Metric & Family & Method & Kodak & Tecnick & DIV2K & CLIC 2020 \\
\midrule
\multirow{11}{*}{LPIPS}
& \multirow{5}{*}{Diffusion} & OneDC & 0.00 & \underline{0.00} & \underline{0.00} & \underline{0.00} \\
& & StableCodec & 8.03 & 32.28 & 27.06 & 36.13 \\
& & AEIC & 41.65 & 34.58 & 25.27 & 32.18 \\
& & DiT-IC & 35.99 & 61.79 & 49.18 & 52.00 \\
& & OSCAR & 127.15 & 225.78 & 201.57 & -- \\
\cmidrule(l){2-7}
& \multirow{6}{*}{VAE-based} & GLC & 21.53 & 47.87 & 33.19 & 39.53 \\
& & DLF & 20.87 & 54.78 & 47.02 & 45.66 \\
& & ProGIC & 36.47 & 56.48 & 42.04 & 84.07 \\
& & EF-LIC & \underline{-3.33} & 15.12 & 3.51 & 24.62 \\
& & RDVQ & 17.06 & 47.25 & 26.35 & 33.11 \\
\cmidrule(l){3-7}
& & \textbf{\ours} & \textbf{-23.19} & \textbf{-11.27} & \textbf{-20.24} & \textbf{-18.39} \\
\midrule
\multirow{11}{*}{DISTS}
& \multirow{5}{*}{Diffusion} & OneDC & 0.00 & 0.00 & 0.00 & 0.00 \\
& & StableCodec & 6.77 & 31.27 & 9.38 & 24.27 \\
& & AEIC & \underline{-16.71} & \underline{-20.02} & \underline{-30.19} & \underline{-21.29} \\
& & DiT-IC & -6.44 & -0.28 & -16.61 & -10.82 \\
& & OSCAR & 131.50 & 240.68 & 223.40 & 339.84 \\
\cmidrule(l){2-7}
& \multirow{6}{*}{VAE-based} & GLC & 54.70 & 72.21 & 50.63 & 70.55 \\
& & DLF & 20.19 & 62.90 & 56.62 & 64.56 \\
& & ProGIC & 80.41 & 126.30 & 78.51 & 172.95 \\
& & EF-LIC & 10.05 & 44.93 & 13.25 & 69.31 \\
& & RDVQ & 9.87 & 27.16 & 9.10 & 27.14 \\
\cmidrule(l){3-7}
& & \textbf{\ours} & \textbf{-37.71} & \textbf{-28.12} & \textbf{-42.46} & \textbf{-31.53} \\
\midrule
\multirow{11}{*}{PSNR}
& \multirow{5}{*}{Diffusion} & OneDC & 0.00 & \textbf{0.00} & 0.00 & \underline{0.00} \\
& & StableCodec & \underline{-17.10} & 41.35 & 31.65 & -- \\
& & AEIC & 2.11 & \underline{1.78} & \textbf{-4.44} & \textbf{-12.35} \\
& & DiT-IC & 33.24 & 50.41 & 40.19 & 26.63 \\
& & OSCAR & 154.09 & 248.22 & 195.61 & 338.47 \\
\cmidrule(l){2-7}
& \multirow{6}{*}{VAE-based} & GLC & 56.04 & 72.28 & 61.07 & 79.51 \\
& & DLF & 42.73 & 78.89 & 63.48 & 67.35 \\
& & ProGIC & 49.84 & 105.02 & 72.35 & 116.69 \\
& & EF-LIC & 11.49 & 61.07 & 39.83 & 59.45 \\
& & RDVQ & 62.13 & 70.23 & 60.92 & 70.88 \\
\cmidrule(l){3-7}
& & \textbf{\ours} & \textbf{-22.30} & 19.60 & \underline{-0.41} & 16.48 \\
\midrule
\multirow{11}{*}{FID}
& \multirow{5}{*}{Diffusion} & OneDC & 0.00 & 0.00 & 0.00 & 0.00 \\
& & StableCodec & \underline{-4.45} & -5.58 & -26.77 & -28.75 \\
& & AEIC & 7.16 & \textbf{-38.95} & \textbf{-49.80} & \textbf{-55.10} \\
& & DiT-IC & 56.47 & 41.17 & 30.56 & 44.53 \\
& & OSCAR & 159.92 & 222.65 & -- & -- \\
\cmidrule(l){2-7}
& \multirow{6}{*}{VAE-based} & GLC & 82.95 & 54.94 & 50.70 & 90.98 \\
& & DLF & 70.80 & 73.87 & 66.43 & 66.97 \\
& & ProGIC & 156.32 & 267.15 & 225.46 & 389.82 \\
& & EF-LIC & 49.01 & 143.61 & 138.95 & 257.90 \\
& & RDVQ & 49.51 & 55.05 & 40.18 & 59.78 \\
\cmidrule(l){3-7}
& & \textbf{\ours} & \textbf{-22.14} & \underline{-32.07} & \underline{-45.68} & \underline{-34.26} \\
\bottomrule
\end{tabular}
\end{table*}

\Cref{tab:bd_selected_appendix} reports OneDC-anchored BD-rate for LPIPS, DISTS, PSNR, and FID on Kodak, Tecnick, DIV2K, and CLIC 2020.
Entries without a valid monotone common interval are left unavailable according to the protocol in \cref{app:protocol}.
CoD-Lite and MS-ILLM are already included in the main Kodak and high-resolution comparisons, so their full entries are not repeated across all four metric blocks in this table.

Across the twelve LPIPS, DISTS, and FID entries, \ours achieves a negative OneDC-relative BD-rate on every dataset.
The corresponding rate and quality curves in
\cref{fig:rd_curves,fig:lightweight_comparison}
show the behavior over the evaluated operating ranges.

The table also places these reconstruction results in the context of model complexity.
AEIC uses approximately $11.2\times$ the parameters and $4.7\times$ the FLOPs of \ours.
StableCodec uses approximately $13.4\times$ the parameters and $14.9\times$ the FLOPs.
EF-LIC and ProGIC use fewer parameters and FLOPs than \ours, while \ours provides lower DISTS and FID over the reported comparisons.

\subsection{High-Resolution Runtime and Memory}

\begin{table*}[t]
\centering
\caption{High-resolution encoding and decoding latency (ms) and peak GPU memory (GiB) on one A100.
Bold and underline mark the two best results.}
\label{tab:highres}
\begingroup
\setlength{\tabcolsep}{3.45pt}
\begin{tabular}{lrrrrrrrrr}
\toprule
\multirow{2}{*}{Method}
  & \multicolumn{3}{c}{1080p}
  & \multicolumn{3}{c}{2K}
  & \multicolumn{3}{c}{4K} \\
\cmidrule(lr){2-4}\cmidrule(lr){5-7}\cmidrule(l){8-10}
  & Enc.
& Dec.
& Mem.
& Enc.
& Dec.
& Mem.
& Enc.
& Dec.
& Mem.
\\
\midrule
ResULIC & 481.75 & 3702.45 & 20.38 & -- & {$>10$ s} & -- & -- & {$>10$ s} & -- \\
StableCodec & 641.76 & 1657.43 & 21.80 & 1085.39 & 3161.54 & 30.65 & -- & {$>10$ s} & -- \\
RDEIC & 683.20 & 2296.58 & 13.55 & 1436.00 & 5963.77 & 20.17 & -- & {$>10$ s} & -- \\
OneDC & 303.33 & 742.06 & 12.81 & 585.72 & 1554.82 & 18.86 & 1542.18 & 4846.63 & 35.38 \\
OSCAR & 513.20 & 1123.38 & 24.75 & 2382.55 & 2719.79 & 79.44 & -- & -- & $>80$ \\
DiT-IC & 480.22 & 750.38 & 10.98 & 718.17 & 1111.59 & 14.47 & 1581.91 & 2448.49 & 25.41 \\
AEIC & 25.15 & 682.50 & 6.95 & \underline{30.69} & 1943.15 & 9.87 & \underline{62.70} & 9303.96 & 17.83 \\
CoD-Lite & 42.37 & 60.47 & 2.45 & 73.61 & 105.09 & 3.97 & 163.43 & 236.05 & 8.33 \\
\midrule
MS-ILLM & 350.85 & 379.01 & 2.99 & 516.47 & 601.18 & 4.87 & 1305.93 & 1613.82 & 9.94 \\
Control-GIC & 610.76 & 2186.30 & 69.99 & -- & -- & $>80$ & -- & -- & $>80$ \\
GLC & 340.746 & 475.738 & 5.66 & 638.989 & 826.327 & 9.71 & 1456.258 & 1816.005 & 20.76 \\
DLF & 915.42 & 1416.49 & 10.00 & 1380.00 & 2153.39 & 12.61 & 3001.08 & 5122.21 & 22.35 \\
ProGIC & 28.40 & 55.00 & 1.07 & 38.63 & 80.38 & 1.46 & 78.26 & 145.24 & 2.58 \\
RDVQ & 704.64 & 809.99 & 6.01 & 953.80 & 1130.17 & 10.08 & 8575.63 & 8655.58 & 20.07 \\
EF-LIC & \underline{19.53} & \underline{35.62} & \underline{0.65} & 31.58 & \underline{55.50} & \underline{1.04} & 65.24 & \underline{116.28} & \underline{2.10} \\
\cmidrule(l){1-10}
\textbf{\ours} & \textbf{14.10} & \textbf{21.44} & \textbf{0.25} & \textbf{22.76} & \textbf{35.95} & \textbf{0.31} & \textbf{49.79} & \textbf{81.14} & \textbf{0.49} \\
\bottomrule
\end{tabular}
\endgroup
\end{table*}

\Cref{tab:highres} extends the runtime comparison to 1080p, 2K, and 4K images on one NVIDIA A100.

The released GLC interface reports analysis and synthesis without a complete bitstream path.
Its forward execution takes
$(159.460,259.762)$ ms at 1080p,
$(291.436,488.817)$ ms at 2K,
and $(659.937,1101.531)$ ms at 4K
for analysis and synthesis, respectively.
Under the complete CompressAI integration defined in \cref{app:protocol}, the corresponding encoding and decoding times are
$(340.746,475.738)$ ms,
$(638.989,826.327)$ ms,
and $(1456.258,1816.005)$ ms.
Peak memory is 5.66, 9.71, and 20.76 GiB at the three resolutions.

Among the available implementations in \cref{tab:highres}, \ours achieves the lowest encoding latency, decoding latency, and peak memory at all three resolutions.
Relative to the second-lowest values, the reductions correspond to factors of
$1.26\times$ to $1.39\times$ for encoding,
$1.43\times$ to $1.66\times$ for decoding,
and $2.58\times$ to $4.27\times$ for peak memory.

EF-LIC provides the second-lowest decoding latency and peak memory across the three resolutions.
AEIC provides the second-lowest encoding latency at 2K and 4K.
These results show that the fast inference of \ours extends from Kodak resolution to high-resolution images.

\subsection{GR Inference Stability}
\label{app:gr_stability}

\paragraph{Inference perturbation.}

We first evaluate the effect of the sampled perturbation at inference using the protocol in \cref{app:gr_controls}.
With the perturbation term removed, the decoder uses only the learned refinement center
$\hat{\bm y}+\bm\delta$.

In the same rate order as \cref{tab:ablation}, the resulting LPIPS and FID pairs are
$(0.262,54.8)$,
$(0.206,36.4)$,
$(0.181,32.5)$,
and $(0.162,27.9)$.

Relative to the complete $\mathrm{M}_4$ results, removing the sampled perturbation changes LPIPS by only 0.000 to 0.002 and increases FID by 0.2 to 0.4.
Most of the gain of the complete GR configuration therefore remains when inference uses only the learned refinement center.

Without sampled perturbation at inference, $\mathrm{M}_4$ retains FID reductions of
7.6, 4.8, 4.0, and 4.0
relative to deterministic $\mathrm{M}_2$.
It also retains reductions of
4.7, 1.5, 1.3, and 1.6
relative to $\mathrm{M}_3$, which uses a channel-wise perturbation scale.
LPIPS remains equal to or lower than both configurations at every evaluated rate.

These results support the interpretation in \cref{sec:ablation} that the adaptive perturbation primarily contributes through training regularization rather than through the particular random realization used at inference.

\paragraph{Decoder seed stability.}

We next vary only the decoder seed according to the protocol in \cref{app:gr_controls}.
Across ten seeds on Kodak, aggregate LPIPS varies by less than $10^{-4}$ and FID by less than $10^{-2}$.

The transmitted indices and payload are identical across these runs.
The small metric variation shows that the reported aggregate reconstruction performance is stable across the evaluated GR realizations.

\paragraph{FID repeatability.}

We also repeat the Kodak FID computation four times with the reconstructed images and patch extraction fixed.
All four runs produce identical FID values.
This confirms that the variation measured above comes from the GR realization rather than repeated execution of the FID estimator.

\section{Additional Visual Comparisons}
\label{app:more_visuals}
\begin{figure*}[t]
  \centering
  \includegraphics[width=\linewidth]{figures/figure_kodak.pdf}
  \caption{Additional Kodak visuals with LPIPS and BPP labels, where lower is better.}
  \label{fig:more_visual_kodak}
\end{figure*}

\begin{figure*}[t]
  \centering
  \includegraphics[width=\linewidth]{figures/figure_tecnick.pdf}
  \caption{Additional Tecnick visuals with LPIPS and BPP labels, where lower is better.}
  \label{fig:more_visual_tecnick}
  \includegraphics[width=\linewidth]{figures/figure_DIV2K.pdf}
  \caption{Additional DIV2K visuals with LPIPS and BPP labels, where lower is better.}
  \label{fig:more_visual_div2k}
  \includegraphics[width=0.96\linewidth]{figures/figure_clic2020.pdf}
  \caption{Additional CLIC 2020 visuals with LPIPS and BPP labels, where lower is better.}
  \label{fig:more_visual_clic}
\end{figure*}

This section provides additional visual comparisons of \ours on all four evaluation datasets.
Each figure places comparisons with high-capacity diffusion codecs above those with lightweight baselines and reports LPIPS and BPP for the displayed reconstructions.
Two further panels extend the Tecnick and CLIC 2020 coverage under the same LPIPS and BPP annotation protocol.

\begin{figure*}[t]
  \centering
  \includegraphics[width=0.96\linewidth]{figures/figure_tecnick_2.pdf}
  \caption{Additional Tecnick visuals with LPIPS and BPP labels, where lower is better.}
  \label{fig:more_visual_tecnick_2}
  \includegraphics[width=0.96\linewidth]{figures/figure_clic2020_2.pdf}
  \caption{Additional CLIC 2020 visuals with LPIPS and BPP labels, where lower is better.}
  \label{fig:more_visual_clic_2}
\end{figure*}

\end{document}